\RequirePackage{fix-cm}
\documentclass[smallextended]{svjour3}       
\smartqed  
\usepackage{epsfig}
\usepackage{helvet}
\usepackage{courier}
\usepackage{floatrow}
\usepackage{multirow}
\newfloatcommand{capbtabbox}{table}[][\FBwidth]
\usepackage{comment}
\usepackage[utf8]{inputenc}
\usepackage[T1]{fontenc}  
\usepackage{url}      
\usepackage{booktabs}     
\usepackage{amsfonts}    
\usepackage{units}      
\usepackage{microtype}
\usepackage{xcolor}
\usepackage{colortbl,booktabs}
\usepackage{soul}
\usepackage{url}
\usepackage[small]{caption}
\usepackage{graphicx}
\usepackage{amsmath}
\usepackage{booktabs}
\usepackage{algorithm}
\usepackage{algorithmic}
\usepackage{epstopdf}
\usepackage{amssymb}
\usepackage{subfigure}
\usepackage{wrapfig}
\usepackage{xspace}
\newcommand{\tabincell}[2]{\begin{tabular}{@{}#1@{}}#2\end{tabular}}

\newcommand{\eg}{\textit{e.g.}\@\xspace}
\newcommand{\wrt}{\textit{w.r.t.}\@\xspace}

\usepackage[flushleft]{threeparttable}

\usepackage[hang,flushmargin]{footmisc}
\begin{document}

\title{Distribution-sensitive Information Retention for Accurate Binary Neural Network}

\author{Haotong Qin \and
        Xiangguo Zhang \and
        Ruihao Gong \and
        Yifu Ding \and
        Yi Xu \and
        Xianglong Liu$^*$
}


\institute{Haotong Qin \at
               State Key Laboratory of Software Development Environment \& Shen Yuan Honors College, Beihang University 
           \and
           Xiangguo Zhang \at
               State Key Laboratory of Software Development Environment, Beihang University 
          \and
          Ruihao Gong \at
               State Key Laboratory of Software Development Environment, Beihang University 
         \and
         Yifu Ding \at
               State Key Laboratory of Software Development Environment \& Shen Yuan Honors College, Beihang University 
        \and
        Yi Xu \at
               State Key Laboratory of Software Development Environment, Beihang University 
        \and
        Xianglong Liu (Corresponding Author) \at
               State Key Laboratory of Software Development Environment, Beihang University \\
              \email{xlliu@buaa.edu.cn}
}

\date{Received: date / Accepted: date}

\maketitle

\begin{abstract}
Model binarization is an effective method of compressing neural networks and accelerating their inference process, which enables state-of-the-art models to run on resource-limited devices.
Recently, advanced binarization methods have been greatly improved by minimizing the quantization error directly in the forward process. However, a significant performance gap still exists between the 1-bit model and the 32-bit one. 
The empirical study shows that binarization causes a great loss of information in the forward and backward propagation which harms the performance of binary neural networks (BNNs).
{We present a novel \textbf{D}istribution-sensitive \textbf{I}nformation \textbf{R}etention \textbf{Net}work (DIR-Net) that retains the information in the forward and backward propagation by improving internal propagation and introducing external representations.
The DIR-Net mainly relies on three technical contributions: (1) \textit{Information Maximized Binarization} (IMB): minimizing the information loss and the binarization error of weights/activations simultaneously by weight balance and standardization; (2) \textit{Distribution-sensitive Two-stage Estimator} (DTE): retaining the information of gradients by distribution-sensitive soft approximation by jointly considering the updating capability and accurate gradient; (3) \textit{Representation-align Binarization-aware Distillation} (RBD): retaining the representation information by distilling the representations between full-precision and binarized networks.}
The DIR-Net investigates both forward and backward processes of BNNs from the unified information perspective, thereby providing new insight into the mechanism of network binarization.
The three techniques in our DIR-Net are versatile and effective and can be applied in various structures to improve BNNs.
{Comprehensive experiments on the image classification and objective detection tasks show that our DIR-Net consistently outperforms the state-of-the-art binarization approaches under mainstream and compact architectures, such as ResNet, VGG, EfficientNet, DARTS, and MobileNet.} 
Additionally, we conduct our DIR-Net on real-world resource-limited devices which achieves $11.1\times$ storage saving and $5.4\times$ speedup.
\end{abstract}

\section{Introduction}
Over the past few years, Artificial Intelligence (AI) utilizing Deep Neural Networks (DNNs), especially Convolutional Neural Networks (CNNs) has shown great potential on some specific tasks such as computer vision, including but not limited to classification~\cite{krizhevsky2012imagenet,VeryDeepConvolutional,7298594,wang2019dynamic,9027877}, detection~\cite{DBLP:journals/corr/GirshickDDM13,DBLP:journals/corr/Girshick15,DBLP:journals/corr/abs-1904-02701,NIPS2015_5638,Li_2019_CVPR} and segmentation~\cite{Everingham:2010:PVO:1747084.1747104,Zhuang_2019_CVPR}. However, deep CNNs usually have a large number of parameters and high computational complexity to satisfy the requirement of high accuracy. Thus a great deal of memory and computing power is always required when running the high accurate CNNs, which significantly limits the deployment of CNNs on lightweight devices such as low-power chips and embedded devices.

\begin{figure*}[t!]
    \begin{center}
        \includegraphics[width=1.0\linewidth]{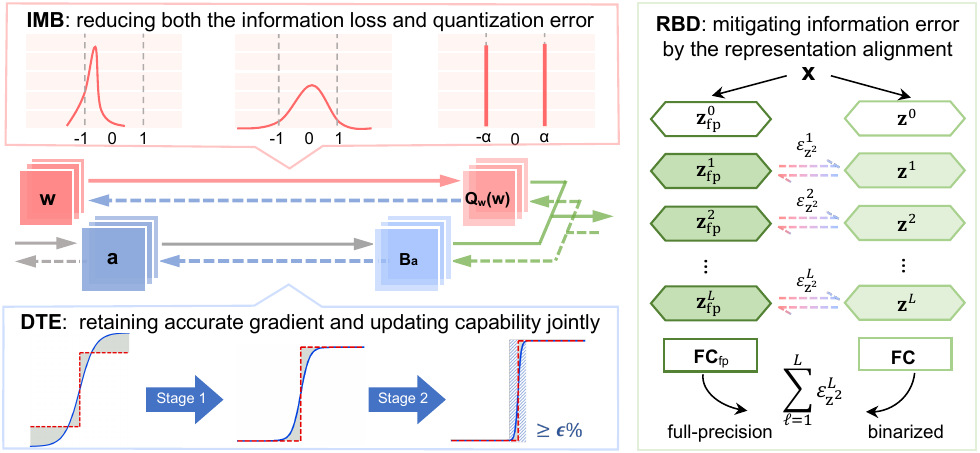}
    \end{center}
    \vspace{-0.2in}
    \caption{Overview of the training and inference process for a convolutional layer of our DIR-Net. {The training consists of Information Maximized Binarization (IMB) in the forward propagation and Distribution-sensitive Two-stage Estimator (DTE) in the backward propagation, and performs under the Representation-align Binarization-aware Distillation (RBD) scheme. IMB changes the weight distribution in the forward propagation to maximizing the information of weights and activations. The shape change of DTE during the whole training process minimizing the loss of gradient information in the backward propagation.
    RDB distills the representations between full-precision and binarized networks.}
    {In the inference process, the convolution operation is implemented by XNOR-Bitcount and Bit-shift operations, which achieve a significant speedup compared with the floating-point convolution.}}
    \label{fig:data_flow}
\end{figure*}

Fortunately, Binary Neural Networks (BNNs) can achieve efficient inference and small memory usage utilizing the high-performance instructions including XNOR, Bitcount, and Shift that most low-power devices support~\cite{Dong_2019_ICCV,Morozov_2019_ICCV,Ajanthan_2019_ICCV,Jung_2019_CVPR,Yang_2019_CVPR,Wang_2019_CVPR,Cao_2019_CVPR,Nagel_2019_ICCV,qin2020bipointnet}. Despite the huge speed advantage, existing binary neural networks still suffer a large drop in accuracy compared with their full-precision counterparts~\cite{DBLP:journals/ijcv/LiuDHZLGD21,DBLP:journals/ijcv/LiuLWYLC20,DBLP:journals/ijcv/SongHGXHS20,DBLP:journals/ijcv/DongNLCSZ19}. The reasons for the accuracy drop mainly lie in two aspects.

On the one hand, the limited representation capability and discreteness of binarized parameters lead to significant information loss in the forward propagation. When the 32-bit parameters are binarized to 1-bit, the diversity of the neural network model drops sharply, which is proved to be the key factor in the accuracy drop of BNNs~\cite{diverse}.
To increase diversity, some work proposed to introduce additional operations. {For example, the ABC-Net~\cite{ABCNet} utilizes multiple binary bases for more representation levels and the WRPN~\cite{mishra2018wrpn} devises wider networks for more parameters.} The Bi-Real Net proposed in \cite{Liu_2018_ECCV} added a full-precision shortcut to the binarized activations to improve the feature diversity, which also greatly improves the BNNs. But due to the speed and memory limit, any extra floating-point calculation or parameter increase will greatly harm the practical deployment on the edge hardware like Raspberry Pi and BeagleBone~\cite{kruger2014benchmarking}. Therefore, it is still a great challenge for BNNs to achieve high accuracy while can be deployed on lightweight devices as well.

On the other hand, accurate gradients supply correct information for network optimization in backward propagation. But during the training process of BNNs, discrete binarization inevitably causes inaccurate gradients and further the wrong optimization direction.
In order to deal with the problem of discreteness, different approximations of binarization for the backward propagation have been proposed~\cite{DBLP:conf/cvpr/CaiHSV17,Liu_2018_ECCV,BNN+,selfBN,ImprovedTraining}, {which can be mainly categorized into improving the updating capability and reducing the mismatching area between the $\operatorname{sign}$ function and the approximate one.}
However, the difference between the early and later training stages is always being ignored. In fact, powerful update capabilities are usually highly required at the beginning of the training process, while small gradient errors become more apparent at the end of training.
Moreover, some works extremely decrease the gap between the $\operatorname{sign}$ function and the estimator in a certain period of the training process, while our study shows that ensuring suitable parameters can be updated in the whole training process is better for BNN optimization.
Specifically, when the estimator of BNN extremely approximates the $\operatorname{sign}$ function, though the gradient error between them is small, the gradient values in BNN are almost all zeros and the BNN can hardly be updated, which is called "saturation"~\cite{Regularize-act-distribution}.
Therefore, the methods devoted to extremely decreasing gradient error may seriously ignore the harm to the parameter updating capability.

In order to address the above-mentioned issues, we study the network binarization from the information flow perspective and propose a novel \textbf{D}istribution-sensitive \textbf{I}nformation \textbf{R}etention \textbf{Net}work (DIR-Net) (see the overview in Fig.~\ref{fig:data_flow}). 
{The proposed DIR-Net mainly relies on three techniques that retain the information during the forward and backward propagation and improve BNNs' training to higher accuracy by improving internal propagation and introducing external representations.}
(1) the DIR-Net introduces a novel binarization approach named \textit{Information Maximized Binarization} (IMB) in the forward propagation, which balances and standardizes the weight distribution before binarizing. With the IMB, we can minimize the information loss in the forward propagation by maximizing the information entropy of the quantized parameters and minimizing the quantization error. Besides, the IMB is conducted offline and thus brings no time cost during inference.
(2) The \textit{Distribution-sensitive Two-stage Estimator} (DTE) is devised to compute gradients in the backward propagation, which minimizes the multi-type information loss by approximating the $\operatorname{sign}$ function. The shape change of the DTE is distribution-sensitive, which obtains the accurate gradients and, more importantly, ensures that there are always enough parameters updated in the whole training process.
{(3) Besides improving internal propagation, a \textit{Representation-align Binarization-aware Distillation} (RBD) is also applied in DIR-Net to improve BNNs' training by introducing external representations. The RBD aligns the forward representations between full-precision and binarized networks by distillation-based optimization to mitigate the information loss of representation caused by binarization.}

Note that we extend our prior conference publication~\cite{IRNet} that mainly concentrates on a binary neural network method. 
This paper further comprehensively studies the information loss of BNN from the perspective of mathematics and experience to comprehend the forward and backward propagation of BNN more deeply. Existing works lack the analysis and comprehension of the information loss in binarization, and the manual or fixed strategies are always applied to BNN but significant information loss still exists.
Therefore, compared with the conference version, this manuscript further comprehensively studies the information loss problem in binarization, presents new distribution-sensitive improvements to the BNN, and compares the proposed method with more SOTA methods on more architectures.
\textbf{First}, we present a more in-depth analysis of the information loss in the forward and backward propagation in Sec.~\ref{DIRNet}. 
{For the forward propagation, we provide mathematical studies about the effect of both the information retention by weight balance and the binarization errors on the global level, which further clarifies the error minimization motivation of our IMB.}
For the backward propagation, we show that the changes in weight distribution during BNN training may limit updating capability of BNNs with soft estimators.
\textbf{Second}, we propose a novel DIR-Net with distribution-sensitive estimator DTE, which improves the backward propagation process. Instead of changing the shape of the estimator in IR-Net~\cite{IRNet} with a fixed strategy, the DIR-Net further adjusts the shape of the estimator according to the distribution of weights/activations in the backward propagation to retain the information of accurate gradients and the updating capability of BNN.
{\textbf{Third}, we design an RBD distillation scheme for DIR-Net to improve BNNs' training by introducing external representations. 
Orthogonal to both the IMB and DTE techniques devoting to retaining information by improving the internal propagation, the RBD aligns the external representations between full-precision and binarized networks by distillation-based optimization to mitigate the information loss of representation caused by binarization.}
\textbf{Fourth}, we add detailed ablation experiments in Sec.~\ref{sec:Exp} to verify the effectiveness of techniques in DIR-Net on BNNs (Table~\ref{ablation_exp}), and also evaluate the impact of binarization errors (Table~\ref{ablation_exp_error}), DTE clipping interval ($t_\epsilon$ setting in Table~\ref{epsilon_exp}), and parameter information entropy (Fig.~\ref{fig:ent_acc}).
{\textbf{Fifth}, we compare proposed DIR-Net on image classification and objective detection tasks with more SOTA binarization approaches in Table~\ref{imagenet} (BONN~\cite{gu2019bayesian}, Si-BNN~\cite{wang2020sparsity}, PCNN~\cite{DBLP:journals/corr/abs-1811-12755}, Real-to-Bin~\cite{martinez2020training}, MeliusNet~\cite{bethge2021meliusnet}, and ReActNet~\cite{liu2020reactnet}), and evaluate it on compact networks (EfficientNet~\cite{tan2019efficientnet}, MobileNet~\cite{mobilenet} and DARTS~\cite{liu2018darts}) and detection frameworks (Faster R-CNN~\cite{ren2015faster} and SSD~\cite{liu2016ssd}).
The results show that our DIR-Net is versatile and effective across various architectures and datasets.}
\textbf{Moreover}, we also add and discuss more latest related work of network compression and quantization~\cite{he2021generative,wang2020towards,phan2020binarizing,chen2020binarized,DBLP:journals/ijcv/LiuLWYLC20,Liu_2019_CVPR,DBLP:journals/ijcv/LiuDHZLGD21,DBLP:journals/ijcv/LiuLWYLC20} to reflect the characteristics and advantages of our DIR-Net.

This work provides a novel and practical view to explain how BNNs work. In addition to the strong capability to retain the information in the forward and backward propagation, DIR-Net has excellent versatility to be extended to various architectures of BNNs and can be trained via the standard training pipeline.
{We evaluate our DIR-Net on image classification (CIFAR-10 and ImageNet) and objective detection tasks (PASCAL VOC and COCO). 
The results indicate that our DIR-Net exceeds other binarization approaches greatly in a variety of architectures, including ResNet, VGG, EfficientNet, MobileNet, DARTS architectures, and Faster R-CNN and SSD detection frameworks.}
To validate the real-world performance of DIR-Net on low-power devices, we also implement it on Raspberry Pi and it achieves outstanding efficiency ($11.1\times$ storage saving and $5.4\times$ speedup).

In summary, our main contributions are listed as follows:
\begin{itemize}
\item {We propose the simple yet efficient Distribution-sensitive Information Retention Network (DIR-Net), which can improve BNNs by retaining information during the training process. Compared with existing fixed-strategy estimators, the DTE estimator in DIR-Net ensures enough updating capability and improves the accuracy of BNNs. And the proposed RBD scheme in DIR-Net aligns the representations of the full-precision network and BNN to mitigate information loss.}
\item We measure the amount of information for binarized parameters by information entropy and present an in-depth analysis about the effects of information loss and binarization error in BNNs.
\item We investigate both forward and backward processes of binary networks from the unified information perspective, which provides new insight into the mechanism of network binarization.
\item {Experiments demonstrate that our method significantly outperforms other state-of-the-art (SOTA) methods in both accuracy and practicality on various architectures and vision tasks. And we further prove the DTE and RBD in the proposed DIR-Net can stably improve the performance.}
\item We implement 1-bit BNNs and evaluate their speed on real-world ARM devices, and the results show that our DIR-Net achieves outstanding efficiency.
\end{itemize}
The rest of this paper is organized as follows. Section II gives a brief review of related model binarization methods and low-power devices. Section III describes the preliminaries of binary neural networks. Section IV describes the proposed approach, formulation, implementation, and discussion in detail. Section V provides the experiments conducted for this work, model analysis, and experimental comparisons with other SOTA methods. In Section VI, we conclude the study.

\section{Related Work}
{Recently, resource-limited embedded devices attract researchers in the area of artificial intelligence by their low-power consumption, tiny size, and high practicality, which significantly promotes the application of artificial intelligence technology.} However, the SOTA neural network models suffer massive parameters and large sizes to achieve good performance in different tasks, which also cause significant complex computation and great resource consumption. 
To compress and accelerate the deep CNNs, many approaches have been proposed, which can be classified into five categories: transferred/compact convolutional filters~\cite{shufflenet,yu2017on,DBLP:journals/pami/WangXXT19}; quantization/binarization~\cite{DBLP:conf/eccv/HuLWZC18,DBLP:conf/nips/ChenWP19,Wu2020Rotation,zhu2019unified}; knowledge distillation~\cite{chen2018darkrank,zagoruyko2017paying,DBLP:journals/pr/DingCH19}; pruning~\cite{han2016deep,he2017channel,ge2017Compressing}; low-rank factorization~\cite{lebedev2015speeding,jaderberg2014speeding,lebedev2016fast,DBLP:journals/pr/WenZXYH18}. 

Compared with other compression methods, model binarization can significantly reduce the consumption of memory. By extremely compressing the bit-width of parameters in neural networks, the convolution filters in binary neural networks can achieve $32\times$ memory saving. Model binarization also makes the compressed model fully compatible with the XNOR-Bitcount operation to achieve great acceleration, and these operations can even achieve $58\times$ speedup in theory~\cite{xnornet}. Besides, the model binarization less changes the architecture compared with other model compression methods, which makes it easier to implement on resource-limited devices and attracts attention from the researchers.
By simply binarizing full-precision parameters including weights and activations, we can achieve obvious inference acceleration and memory saving.

\cite{DBLP:journals/corr/CourbariauxB16} proposed a binarized neural network by simply binarizing the weight and activation to +1 or -1, which compressed the parameters and accelerated CNNs by efficient bitwise operations.
However, the binarization operation in this work caused a significant accuracy drop.
After this work, many binarization approaches were designed to decrease the gap between BNNs and full-precision CNNs.
{The XNOR-Net~\cite{xnornet} is one of the most classic model binarization methods, which pointed out that using floating-point scalars for each binary filter can achieve significant performance improvement.} 
Therefore, it proposed a deterministic binarization method which reduces the quantization errors of the output matrix by applying the 32-bit scalars in each layer, while it incurred more resource-consuming floating-point multiplication and addition.
{The TWN~\cite{DBLP:journals/corr/LiL16} and TTQ~\cite{DBLP:journals/corr/ZhuHMD16} utilized more quantization points to improve the representation capability of quantized neural networks.} 
{Unfortunately, the bitwise operation can never be used in these methods to accelerate the network, and the memory consumption also increased.}
The ABC-Net~\cite{ABCNet} shown that approximating weights and activations by applying multiple binary bases can greatly improve the accuracy of BNNs, while it unavoidably decreases the compression and acceleration ratios. The HWGQ~\cite{DBLP:conf/cvpr/CaiHSV17} considered the quantization error from the perspective of activation function. The LQ-Nets~\cite{LQ-Net} applied a large number of learnable full-precision parameters to get better performance while increasing the memory usage.
\cite{martinez2020training} got strong BNNs with a multi-step training pipeline and a well-designed objective function in the training process.
Some binarization methods are devoted to solving the gradient error caused by approximating the binarization ($\operatorname{sign}$) function by a well-designed estimator in the backward propagation. 
BNN+~\cite{BNN+} also proposed an estimator to reduce this gap and further studied various estimators to find a better solution. DSQ~\cite{DSQ} and IR-Net~\cite{IRNet} creatively applied soft estimators that gradually changes its shape to optimize the network.
Bi-Real~\cite{Liu_2018_ECCV} introduced a novel BNN-friendly architecture with Bi-Real shortcut to improve the performance from the term of accuracy.
{ReActNet~\cite{liu2020reactnet} further improved the architecture and training steps and achieved a better BNN performance. ReActNet (1) applies RSign and RPReLU to enable explicit learning of the distribution to reshape and shift the values around zero, (2) uses the ReActNet block structure that duplicates input channels and concatenates two blocks with the same inputs to address the channel number difference, (3) uses average pooling in the shortcut to match spatial downsampling.} 

Though some progress has been made on model binarization, existing binarization approaches still cause a serious decrease in accuracy compared with 32-bit models.
{First, since the existing works do not effectively measure and retain the information in BNNs, the  massive information loss is still a severe problem exists in the BNN training process. Second, the existing methods only focus on minimizing the gradient error, and seriously neglect the update capability of network parameters. It is a trade-off between update capability and accurate gradient that researchers should take into account when designing the estimators.}
Additionally, the existing methods, which were proposed to increase the accuracy of BNNs, always incur extra floating-point multiplication or addition.
Thus we propose DIR-Net to retain the information during the training process of BNNs. Further, it eliminate the resource-consuming floating-point operations in the convolutional layer.

\section{Preliminaries}

The main operation in a layer of DNNs in the forward propagation can be expressed as
\begin{equation}
z=\mathbf w\otimes \mathbf a ,
\end{equation}
where $\otimes$ indicates the inner product operation, $\mathbf w \in \mathbb R^n$ and $\mathbf a\in\mathbb R^n$ represent weight tensors and the input activation tensors, respectively. $\mathbf a\in\mathbb R^n$ is the output of the previous layer. However, a large number of floating-point multiplications greatly consume memory and computing resources, which heavily limits the applications of CNNs on embedded devices. 

Previous work has shown that bitwise operations, including XNOR, Bitcount, and Shift, can greatly accelerate the inference of CNNs on low-power devices~\cite{xnornet}.
Therefore, in order to compress and accelerate the deep CNNs, binary neural networks binarize the 32-bit weights and/or activations to 1-bit. In most cases, binarization can be expressed as
\begin{equation}
\label{eq:quantized}
\mathbf{Q_x}=\alpha\mathbf{B_x},
\end{equation}
where $\mathbf x$ indicates 32-bit weights $\mathbf w$ or 32-bit activations $\mathbf a$, and $\mathbf B_{\mathbf x}\in{\{-1, +1\}}^n$ represents  binary weights $\mathbf B_{\mathbf w}$ or binary activations $\mathbf B_{\mathbf a}$. $\alpha$ represents scalars including ${\alpha}_{\text{w}}$ for binary weights and ${\alpha}_{\text{a}}$ for binary activations. Usually, the $\operatorname{sign}$ function is used to calculate $\mathbf B_{\mathbf x}$
\begin{equation}
\label{eq:binarized}
\mathbf{B_x}=\operatorname{sign}(\mathbf{x})=
\begin{cases}
+1,& \mathrm{if} \ \mathbf x \ge 0\\
-1,& \mathrm{otherwise}.
\end{cases}
\end{equation}

With the binary weights and activations, the tensor multiplication operations can be approximated by
\begin{equation}
z = \mathbf{Q_w}\otimes \mathbf{Q_a}= {{\alpha}_{\text{w}}}{{\alpha}_{\text{a}}}({\mathbf{B_w}}\odot {\mathbf{B_a}}),
\end{equation} 
where $\odot$ indicates the bitwise inner product operation of tensors implemented by bitwise operations XNOR and Bitcount. In addition, since the Shift operation is more hardware-friendly, some work even replace the multiplication in the inference process of BNNs by Shift, such as the Shift-based batch normalization~\cite{DBLP:journals/corr/CourbariauxB16}, which further accelerates the inference speed of BNNs on hardware.

However, the derivative of the $\operatorname{sign}$ function is zero almost everywhere, which is obviously incompatible with the backward propagation since exact gradients for activations and/or weights before the discretization would be zero.
Therefore, many works adopt the Straight-Through Estimator (STE) \cite{bengio2013estimating} in gradient propagation, which is $\operatorname{identity}$ or $\operatorname{hardtanh}$ function specifically.

\section{Distribution-sensitive Information Retention Network}
\label{DIRNet}

In this paper, we mention that severe information loss during training hinders the high accuracy of BNNs. To be exact, information loss is mostly caused by the $\operatorname{sign}$ function in the forward propagation and the approximation of gradients in the backward propagation, and it greatly limits the performance of BNNs. To address this problem, we propose a novel network, \textbf{D}istribution-sensitive \textbf{I}nformation \textbf{R}etention \textbf{Net}work (DIR-Net), which retains information during training and deliver excellent performance to BNNs. Besides, all convolution operations in DIR-Net are replaced by hardware-friendly bitwise operations.

\subsection{Information Maximized Binarization in the Forward Propagation}
\label{sec:IMB}
In the forward propagation, the BNN usually suffers both information entropy decreases and quantization error, which further causes information loss of weights and activations. To retain the information and minimize the loss in the forward propagation, we propose Information Maximized Binarization (IMB) that jointly considers both information loss and quantization error.

\subsubsection{Information Loss in the Forward Propagation}
Since the discretization of the parameters by the binarization operation, the full-precision and binarized parameters suffer a large numerical difference causing significant information loss.
In order to make the parameters of the binarized network closer to the full-precision counterparts, the binarization error of the BNN should be minimized.
Consider the computation in a multivariate function $h(\mathbf x)$, where $\mathbf x$ denotes the variable vector with full-precision. 
When the $h(\cdot)$ function represents a neural network, $\mathbf x$ represents the 32-bit parameters (weights $\mathbf w$/activations $\mathbf a$).
The global error $\mathcal{E}_h$ caused by quantizing $\mathbf x$ can be expressed as
\begin{equation}
\mathcal{E}_h=h(\mathbf {Q_x})-h(\mathbf x),
\end{equation}
where $\mathbf{Q_x}$ indicates the variable vector quantized from $\mathbf x$. 
When the probability distribution of $\mathbf x$ is known, the error distribution and the moments of the error can be computed. For example, the minimization of expected absolute error can be present as
\begin{equation}
\label{eq:exp_error}
\min \mathbb E\left[\left|\mathcal{E}_{h}\right|\right]= 
\int \lvert h\left(\mathbf{Q_x}\right) 
-h\left(\mathbf{x}\right) \rvert f(\mathbf x)\ d \mathbf x,
\end{equation}
where $\mathbb E[\cdot]$ denotes the expectation operator and $f(\mathbf x)$ denotes the probability density function of $\mathbf x$.
In general, $h(\cdot)$ can be any linear or nonlinear function of its arguments, and an analytical evaluation of this multidimensional integral can be very difficult. 
In prior work~\cite{93812,35496}, a simplifying assumption is made where the quantity of $\mathcal{E}_h$ is approximated by its first-order Taylor series expansion
\begin{equation}
\mathcal{E}_{h}\approx\frac{\partial h}{\partial \mathbf x} \left(\mathbf {Q_x}-\mathbf x\right).
\end{equation}
For a certain value of $\mathbf x$, the $\frac{\partial h}{\partial \mathbf x}$ is constant and non-zero.
Therefore, minimizing the global error $\mathbb E\left[\left|\mathcal{E}_{h}\right|\right]$ can be approximated as minimizing the quantization error between the quantized (binarized) vectors and the full-precision counterparts. The optimization problem in Eq.~(\ref{eq:exp_error}) can be simplified as
\begin{equation}\label{quantizatoin_error}
\min J(\mathbf{Q_x})=\lvert{\mathbf{Q_x}-\mathbf x}\rvert ,
\end{equation}
where $J(\mathbf{Q_x})$ is the quantization error of quantized parameters.

There are many studies, such as~\cite{xnornet,DBLP:journals/corr/abs-1708-08687,LQ-Net}, that focus on binarized neural networks, optimizing the quantizer by minimizing the quantization error.
Their objective functions (Eq.~(\ref{quantizatoin_error}) typically) suppose that quantized models should just strictly follow the pattern of full-precision models, which is not always enough, especially when the parameters are quantized to extremely low bit width. 
{And during the BNN training, the parameters quantized from the full-precision counterparts are limited to 2 values by the binarization function. Thus, the amount of parameter information estimated by Shannon entropy descends from almost infinite (32-bit) to extremely limited (1-bit), and is vulnerable since the parameters suffer highly homogenization.}
{Besides, the representation space of binary neural networks $\mathbf{B_x}\in \{-1, 1\}^{n\times n}$ is also quite different from that of full-precision neural networks $\mathbf{X}\in \mathbb{R}^{n\times n}$.}
Without retaining the information during training, it is insufficient and difficult to ensure a highly accurate binarized network only by minimizing the quantization error.

Therefore, our study is basically derived from the perspective of information retention. 
{In BNNs, the binarized parameter $\mathbf{X}$ optimized to retain information should most reflect the original full-precision counterparts $\mathbf{Q_x}$. 
And in information theory, this goal can be expressed as maximizing the mutual information $\mathcal{I}(\mathbf{X}; \mathbf{Q_x})$ between the full-precision and binarized parameters:}
\begin{equation}
\label{eq:im}
\arg\max\limits_{\mathbf{X}, \mathbf{Q_x}}\ \mathcal{I}(\mathbf{X} ; \mathbf{Q_x})
= \mathcal{H}(\mathbf{Q_x}) - \mathcal{H}(\mathbf{Q_x}\mid \mathbf{X}),
\end{equation}
{where ${\mathcal{H} (\mathbf{Q_x})}$ is the information entropy, and ${\mathcal {H} (\mathbf{Q_x}\mid \mathbf{X})}$ is the conditional entropy of $\mathbf{Q_x}$ given $\mathbf{X}$. Since we use the deterministic $sign$ function as the quantizer in the binarization, for any $\mathbf{X}$ there is one and only one corresponding $\mathbf{X}$, \textit{i.e.}, $\mathcal{H}(\mathbf{Q_x}\mid \mathbf{X})=0$.
Hence, the original objective function Eq.~(\ref{eq:im}) is equivalent to maximizing the information entropy of $\mathbf{Q_x}$.}

Then we state the precise definition of the information entropy in BNNs and then analyze how to maximize it.
For a random variable $b\in\{-1, +1\}$ obeying Bernoulli distribution, each element in $\mathbf{B_x}$ can be viewed as a sample of $b$.
The information entropy of $\mathbf{Q_x}$ in Eq.~(\ref{eq:quantized}) can be calculated by
\begin{equation}\label{ie}
\mathcal{H}(\mathbf{Q_x})=\mathcal{H}(\mathbf{B_x})=-\sum\limits_{b\in\{-1,1\}}p(b)\ln\left(p(b)\right),
\end{equation}
where $p(b)$ denote the probability, $p(b)\in[0,1]$ and $p(1)+p(-1)=1$.
By maximizing the information entropy $\mathcal{H}(\mathbf{Q_x})$ in Eq.~(\ref{ie}), we make the binarized parameters $\mathbf{Q_x}$ have the maximized amount of information, so that the information in the full-precision counterpart $\mathbf x$ is retained.

\subsubsection{Information Retention via Information Maximized Binarization}

To retain the information and minimize the loss in the forward propagation, we propose Information Maximized Binarization (IMB) that jointly considers both information loss and quantization error.
First, we balance weights of the BNN to maximize the information of weights and activations.
Under the Bernoulli distribution assumption and symmetric assumption of $\mathbf x$, when $p(1)=0.5$ in Eq.~(\ref{ie}), the information entropy $\mathcal H(\mathbf{Q_x})$ of the quantized values $\mathbf x$ takes the maximum value, which means the binarized values should be evenly distributed. However, it is non-trivial to make the weight of BNNs be close to that uniform distribution only through backward propagation.

\begin{wrapfigure}[14]{r}{0.5\textwidth}
\vspace{-0.2in}
\includegraphics[width=0.99\textwidth]{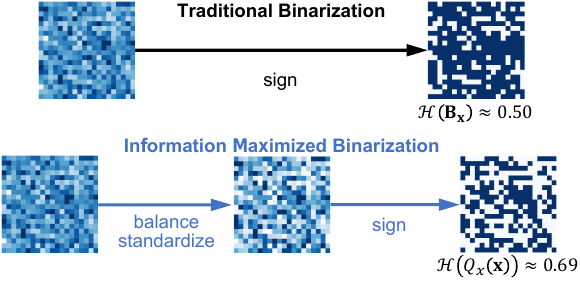}
\vspace{-0.3in}
\caption{Comparison on information entropy of binary weights quantized with IMB $\mathbf{Q_x}$ and the $\operatorname{sign}$ function, respectively. Owing to the balance characteristic brought by IMB, the information entropy of $\mathbf{Q_x}$ is larger than $\operatorname{sign}(\mathbf{x})$, where $\mathbf{Q_x}$ and $\operatorname{sign}(\mathbf{x})$ have a probability of 0.5 and 0.2 to take value 1 under Bernoulli distribution, respectively.}
\label{fig:Libra-PB}
\end{wrapfigure}

Fortunately, we find that simply redistribute the full-precision counterpart of binarized weights can maximize the information entropy of binarized weights and activations simultaneously.
Our IMB balances weights to have zero-mean attribute by subtracting the mean of full-precision weights.
{Moreover, directly binarizing weights without performing the unit-norm might cause two problems. First, the full-precision weights are usually heavily clustered in the range $[-1, 1]$, which leads to more significant gradient approximation errors using STE. For example, in the full-precision ResNet-18 model for ImageNet, the proportion of weight elements in the range $[-1, 1]$ can even reach more than 95\%. Second, since the sign function with zero threshold is applied for the binarization, a large number of weights close to 0 would change the signs and the binarized values due to tiny gradient perturbations, which can make the optimization of BNNs severely unstable.}
Thus, we further standardize the balanced weights to mitigate the negative effect of weight magnitude. 
The standardized balanced weights $\hat{\mathbf w}_{\text{std}}$ are obtained through standardization and balance operations as follows
\begin{equation}\label{eq:redistribute}
\hat{\mathbf w}_{\text{std}}={\mathbf w_{\text{std}}}-\mu(\mathbf w_{\text{std}}), \quad \mathbf w_{\text{std}}=\frac{\mathbf w}{\sigma({\mathbf w})},
\end{equation}
where $\mu$ and $\sigma$ denote the mean and standard deviation, respectively. $\hat{\mathbf w}_{\text{std}}$ has two characteristics: (1) $zero\verb|-|mean$, which maximizes the obtained binary weights' information entropy. 
(2) $unit\verb|-|norm$, {which avoids the full-precision counterparts of binarized weights too concentrated.}
{Therefore, compared with the direct use of the balanced progress, the use of standardized balanced progress makes the weights in the network stably updated and thus further improves the performance of BNNs.}

Since the value of $\mathbf{Q_w}$ depends on the sign of $\hat{\mathbf w}_{\text{std}}$ and the distribution of $\mathbf{w}$ is almost symmetric~\cite{Simultaneously-Optimizing-Weight,ACIQ}, 
the balanced operation can maximize the information entropy of quantized $\mathbf{Q_w}$ on the whole.
And when IMB is used for weights, the information flow of activations in the network can also be maintained. Supposing quantized activations $\mathbf{Q_a}$ have mean $\mathbb{E}[\mathbf{Q_a}] = \mu\mathbf 1$, the mean of $\textbf{z}$ can be calculated by
\begin{equation}
\mathbb{E}[\textbf{z}] = {\mathbf{Q_w}}\otimes\mathbb{E}[\mathbf{Q_a}] = {\mathbf{Q_w}}\otimes\mu\mathbf 1.
\end{equation}
Since the IMB for weights is applied in each layer, we have ${\mathbf{Q_w}}\otimes\mathbf 1 = 0$, and the mean of output is zero. Therefore, the information entropy of activations in each layer can be maximized, which means that the information in activations can be retained.

Then, to further minimize the quantization error defined in Eq.~(\ref{quantizatoin_error}) and avoid extra expensive floating-point calculations in previous binarization methods causing by 32-bit scalars, the IMB introduces an integer shift-based scalar $s$ to expand the representation capability of binary weights. The optimal shift-based scalar can be solved by
\begin{equation}
\mathbf{B_{w}^*},s^*=\mathop{\arg\!\min}\limits_{\mathbf{B_{w}}, s}\lVert{\hat{\mathbf w}_{\text{std}}-\mathbf{B_{w}}\ll\gg s}\rVert^2\quad s.t.\  s\in\mathbb N,
\end{equation}
where $\ll\gg$ stands for left or right Bit-shift. $\mathbf{B_{w}^*}$ is calculated by $\mathbf{B_{w}^*}=\operatorname{sign}(\mathbf{\hat w}_{\text{std}})$, thus $s^*$ can be solved as
\begin{equation}
    s^*=\operatorname{round}(\log_2 {\frac{||\hat{\mathbf w}_{\text{std}}||_{1}}{n}}).
\end{equation}

Therefore, our IMB for the forward propagation can be presented as below:
\begin{equation}
\label{eq:Libra-PB}
\begin{aligned}
\mathbf{Q_w}=\mathbf{B_{w}}{\ll\gg}{s}&=\operatorname{sign}(\hat{\mathbf w}_{\text{std}}){\ll\gg}{s},\\
\mathbf{Q_a}=\mathbf{B_{a}}&=\operatorname{sign}(\mathbf a).
\end{aligned}
\end{equation}
The main operations in DIR-Net can be expressed as
\begin{equation}
z=({\mathbf{B_{w}}}\odot {\mathbf{B_{a}}})\ll\gg s.
\end{equation}

As shown in Fig.~\ref{fig:Libra-PB}, the parameters quantized by IMB have the maximum information entropy under the Bernoulli distribution. We call our binarization method "Information Maximized Binarization" because the parameters are balanced before the binarization operations to retain information.

Note that IMB serves as an implicit rectifier that reshapes the data distribution before binarization. In the literature, a few studies also realized this positive effect on the performance of BNNs and adopted empirical settings to redistribute parameters~\cite{xnornet,Regularize-act-distribution}. For example, \cite{Regularize-act-distribution} proposed the specific degeneration problem of binarization and solved it using a specially designed additional regularization loss. Different from these work, we first straightforwardly take the information perspective to rethink the impact of parameter distribution before binarization and provide the optimal solution by maximizing the information entropy. In this framework, IMB can accomplish the distribution adjustment by simply balancing and standardizing the weights before the binarization. This means that our method can be easily and widely applied to various neural network architectures and be directly plugged into the standard training pipeline with a very limited extra computation cost. Moreover, since the convolution operations in our DIR-Net are thoroughly replaced by bitwise operations, including XNOR, Bitcount, and Shift, the implementation of DIR-Net can achieve extremely high inference acceleration on edge devices.

\begin{figure}[tbp]
\includegraphics[width=0.8\textwidth]{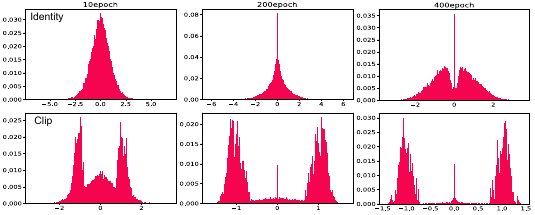}
\vspace{-0.1in}
\caption{{The weight distributions of a specific convolutional layer in binarized ResNet-18 architecture with IMB on ImageNet dataset at different epochs (10, 200, and 400) during training.} The first and second rows present the distributions in the BNNs using $\operatorname{identity}$ and $\operatorname{clip}$ approximations, respectively.}
\label{fig:trend}
\end{figure}

\subsection{Distribution-sensitive Two-stage Estimator in the Backward Propagation}
\label{sec:dte}

In the backward propagation, affected by the limited update range of the estimator and the gradient approximation error simultaneously, the gradient of the BNN suffers from information loss. 
In order to retain the information originated from the loss function in the backward propagation, we propose a progressive Distribution-sensitive Two-stage Estimator (DTE) to obtain the approximation of gradients.

\subsubsection{Information Loss in the Backward Propagation}

Due to the discretization caused by binarization, the approximation of gradients is inevitable in the backward propagation. Therefore, since the impact of quantization cannot be accurately modeled by approximation, a huge loss of information occurs. The approximation can be formulated as
\begin{equation}
\frac{\partial \mathcal L}{\partial\mathbf w}=\frac{\partial \mathcal L}{\partial \mathbf{Q_w}}\ \frac{\partial \mathbf{Q_w}}{\partial \mathbf{w}} \approx \frac{\partial \mathcal L}{\partial \mathbf{Q_w}}\ g'(\mathbf w),
\end{equation}
where $L(\mathbf w)$ indicates the loss function, $g(\mathbf w)$ represents the approximation of the $\operatorname{sign}$ function and $g'(\mathbf w)$ donates the derivative of $g(\mathbf w)$. In previous work, there are two commonly used approximations practices
\begin{equation}
    \operatorname{identity}: y=x \quad \text{or} \quad
    \operatorname{clip}: y=\operatorname{hardtanh}(x).
\end{equation}  

The $\operatorname{identity}$ function completely ignores the effect of binarization and directly passes the gradient information of output values to input values. As shown in the shaded area of Fig.~\ref{fig:evolution}(a), the gradient error is huge and accumulates through layers during the backward propagation. In order to avoid unstable training instead of ignoring the error caused by $\operatorname{identity}$, it is necessary to design a better estimator to retain accurate information of gradient.

The $\operatorname{clip}$ function considers the clipping attribute of binarization, which means only those inside the clipping interval ($[-1, 1]$) can be passed through backward propagation. But only the gradient information inside the clipping interval can be passed. As shown in Fig.~\ref{fig:evolution}(b), as for parameters outside $[-1, 1]$, the gradients are clamped to zero, which means that once the value jumps outside of the clipping interval, it will not be updated anymore. 
This feature greatly limits the updating capability of backward propagation, thereby the $\operatorname{clip}$ approximation makes optimization more difficult and harms the accuracy of models. 
Strong updating capability is essential for the training of BNNs, especially at the beginning of the training process.

Existing estimators are designed to obtain the gradient close to the derivative of the sign function and retain the updateable capability of the BNN, so most of them have an updateable interval, \eg, for the clip function, the interval is $[-1,1]$. 
However, we observe an interesting trend during the training process about the changes in the distribution of weights. 
As Fig.~\ref{fig:trend} shown, when the IMB is applied, the number of weights close to 0 continuously decreases during training, which occurs in most BNNs with various estimators (such as $\operatorname{identity}$ and $\operatorname{clip}$ approximation).
{Moreover, since the application of batch normalization in BNN, there are also a large number of elements in the activation excluded from the updateable interval, such as $[-1, 1]$ for STE.}
The phenomenon causes more weights to be outside the updateable interval and brings great challenges to the design of estimators.
For BNNs with $\operatorname{clip}$ approximation, the phenomenon lets more weights be out of $[-1,1]$ and these can not be updated anymore, which limits their updating capability seriously.
Some soft approximation functions designed to reduce the gradient error are also affected by this problem since they reduce the updateable interval of the parameter as well.
For example, in the later stage of the EDE in IR-Net~\cite{IRNet}, the update range of the estimator continues to shrink to reduce the information loss caused by gradient errors. At the end of this stage, less than 3\% of the weights can be updated (Fig.~\ref{fig:DTE_w_d}). In other words, the BNN almost lost its updating capability at this time.

The $\operatorname{identity}$ function causes gradient error between the $\operatorname{sign}$ function in binarization and the gradients in backward propagation, while the $\operatorname{clip}$ and soft approximation functions cause part of gradient outside the updatable interval. 
Our method try to make a trade-off to take the advantage of these two types of gradient approximation and avoid being affected by their drawbacks.

\subsubsection{Information Retention via Distribution-sensitive Two-stage Estimator}

\begin{figure*}[tp]
\centering
\includegraphics[width=1.0\linewidth]{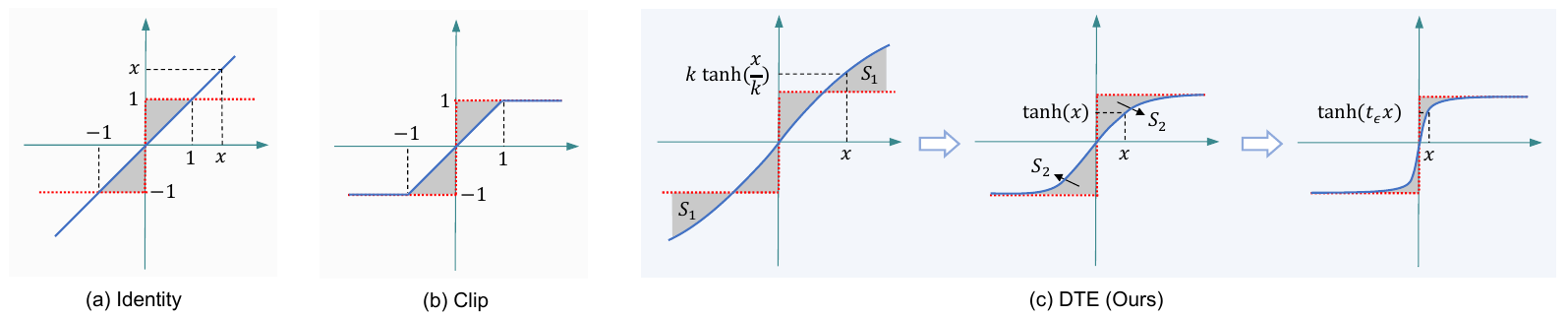}
\vspace{-0.2in}
\caption{Error caused by gradient approximation, represented by the area of gray shades. As is shown, (a) $\operatorname{identity}$ approximation suffers huge error. (b) $\operatorname{Clip}$ approximation does not update the values outside the clipping interval. {(c) DTE maintains the updating capability at early stage and progressively reduces the error.} $S_1$ shrinks during Stage 1 by decreasing the clipping value and $S_2$ shrinks during Stage 2 by increasing the derivate.}
\label{fig:evolution}
\end{figure*}

To make a balance between them, and obtain the optimal approximation of gradients in the backward propagation, we proposed Distribution-sensitive Two-stage Estimator (DTE).
{Based on the analysis above, an intuitive backward propagation solution is using a shape-flexible soft function to approximate the gradient for meeting the information retention requirements in different training stages. The hyperbolic tangent (tanh) function $y=\tanh(x)$ is one of the ideal basic approximation functions, which has a centrosymmetric shape same as the sign function, and also has limits of 1 and -1 when $x\rightarrow\infty$ and $x\rightarrow-\infty$, respectively.
And in order to make the function more flexible to adapt to the requirements of different stages, the form of DTE is defined as}
\begin{equation}
    g(x) = k\tanh{tx} ,
\end{equation}
{where $k$ and $t$ denote the coefficients that determine the specific shape of the function,} $g(x)$ represents the derivable approximate substitute for the forward $\operatorname{sign}$ function in the backward propagation, and $x$ denotes the random variable sampled from the full-precision parameter $\mathbf x$.
The $k$ and $t$ are distribution-sensitively change during the training process to restrict the shape of approximate function:
\begin{equation}
\label{eq:DTE}
\begin{aligned}
t = \min({t_{100\%}}, \max(T_{\min}&10^{\frac{i}{N}\times\log \frac{T_{\max}}{T_{\min}}}, t_{\epsilon})), \quad k = \max(\frac{1}{t}, 1),\\
\mathit{s.t. } &\sum\limits_{-t\le i\le t}p(i)>\epsilon,
\end{aligned}
\end{equation}
where $i$ denotes the current epoch, $N$ is the number of total epochs; $t_{\epsilon}$ means that the number of parameters in the range $[-t, t]$ is $\epsilon$ of the total, where $\epsilon$ indicates the lower limit for the percentage of parameters with high updating capability and is empirically set as $10\%$ to take both updating capability and accurate gradient into account; {$t_{100\%}$ is defined as $t_{100\%}=\max(|\textbf{x}|)$, which determines the upper limit of valid update range for DTE;} $p(\cdot)$ is probability mass function of $x$ that reflects the distribution of the element values in the parameter $\mathbf x$; $T_{\min}$ and $T_{\max}$ are $10^{-1}$ and $10^1$, respectively.

In order to retain the information originated from the loss function in the backward propagation, the DTE proposes a progressive distribution-sensitive two-stage method to obtain the approximation of gradients. 

\noindent\textbf{Stage 1: {Retain the updating capability of the backward propagation algorithm.}} We keep the derivative value of gradient estimation function near one, and then gradually reduce the clipping value from a larger number to one. {At the start of this stage, the shape of DTE is depending on the weight distribution of each layer, which ensures all parameters to be fully updated. DTE adaptively changes the clipping value during this stage to get more accurate gradients.}
{The derivation of the DTE in the first stage is presented as:}
\begin{equation}
\label{eq:DTE_s1}
{g'(x)=1-\mathrm{tanh}^2(tx), \ t > 1.}
\end{equation}
Applying this method, our estimation function evolves from $\operatorname{identity}$ to $\operatorname{clip}$ approximation, {which ensures the strong updating capability at the beginning of the training process and alleviates the loss of updating capability.}

\noindent\textbf{Stage 2: {Keep the balance between accurate gradients and strong updating capability.}} In this stage, we keep the clipping value as one and gradually push the derivative curve towards the shape of the step function, and ensure that enough parameters are updated during this process.
During this process, the shape of DTE is changed according to the parameter distribution, and the derivative around 0 is continuously increased to obtain an accurate gradient until there are not enough parameters to be updated.
The derivation of the DTE in the second stage is presented as:
\begin{equation}
\label{eq:DTE_s2}
g'(x)=t\cdot (1-\mathrm{tanh}^2(tx)), \ t_{\epsilon} \le t \le 1.
\end{equation}

Benefited from the proposed method, our estimation function evolves from $\operatorname{clip}$ approximation to the $\operatorname{sign}$ function, which ensures the consistency in forward and backward propagation.

Fig.~\ref{fig:evolution}(c) shows the shape change of DTE in each stage. Our DTE updates all parameters in the first stage, and further improves the accuracy of parameters in the second stage. Based on this two-stage estimation, DTE can reduce the gap between the forward binarization function and the backward approximation function. Meanwhile, the shape of DTE is adaptively adjusted by parameter distribution to ensure that a certain volume of parameters can be updated in each iteration. And in this way, all the parameters can be reasonably updated.

\subsection{{Representation-align Binarization-aware Distillation Scheme for Information Retention Training}}

{Although the forward and backward propagation of BNNs is greatly improved to reduce the information loss, the high discretization still affects the representations inevitably.
With well-trained corresponding full-precision networks, we present a Representation-align Binarization-aware Distillation (RBD) scheme to improve BNNs orthogonally with the improved internal propagation by introducing external representations.}

\subsubsection{{Information Loss in the Internal Representation}}

{In BNNs, the parameters are binarized from the corresponding full-precision counterparts during the training process. Our proposed binarization unit is a deterministic binarization strategy that minimizes numerical errors and entropy by weight balance and standardization, aiming to minimize the impact of binarization on the information flow of BNNs.
These techniques improve the BNN performance by improving its own parameters.
However, the gain of improving the internal propagation purely is limited, thus the information loss in the BNNs still exists and is even significant.
We can also optimize the representation of the BNNs by introducing the assistance of external information, such as applying the representation of the full precision model as a reference.}

{For the convolutional layer in the BNN, the error jointly caused by activation and weight binarization on the output representation is still hard to estimate. Specifically, consider a binary neural network $M$ and its full-precision counterpart $M_\text{fp}$, the error $\mathcal{E}_\text{r}$ of output representation can be expressed as}
\begin{equation}
\label{eq:output_error}
\mathcal{E}_\textbf{z}=\frac{\textbf{z}_\text{fp}}{\|\textbf{z}_\text{fp}\|_{2}}-\frac{\textbf{z}}{\|\textbf{z}\|_{2}},
\end{equation}
{where $\textbf{z}_\text{fp}=\mathbf{W} \otimes \mathbf{A}; \textbf{W}, \textbf{A} \in M_\text{fp}$ and $\textbf{z}=\mathbf{Q_w} \odot \mathbf{Q_a}; \mathbf{Q_w}, \mathbf{Q_a} \in M$. Considering the batch normalization layer usually followed by the convolutional layer, we normalize the representation by L2 normalization function $\|\cdot\|_2$ to get the rid of the interference of scale.
Due to the nonlinearity of the convolution and binarization operations in Eq.~(\ref{eq:output_error}), it is hard to directly solve or estimate the error by statistic-based methods.
Moreover, when the binarization is considered as perturbation the network, the errors caused by quantization may be continuously accumulated in the BNN~\cite{lin2018defensive}.}

{We would like to note that the above discussion about the information loss of internal representation in the BNN is compatible with the related conclusion for the binarization unit in Sec.~\ref{sec:IMB} but cannot be mitigated fundamentally by the binarization unit improvements. 
The essence of information retention in Eq.~(\ref{eq:exp_error}) is to minimize the loss of information caused by a binarization unit, which regards a binarized parameter (such as the weight and activation of a certain layer) as the variable and minimizes the global error caused by binarization.
However, when we regard more than one parameter as variables, it is hard to directly estimate or optimize the joint effect of their binarization on a global error by a solution-based approach.
Besides, there is also a significant accumulated error in the information flow in the forward propagation of the BNN, especially in the latter part of the propagation.}

\subsubsection{{Information Retention via Representation-align Binarization-aware Distillation}}

{Therefore, except the techniques to improve the binarization unit, we also present an Representation-align Binarization-aware Distillation (RBD) scheme for the BNN training by introducing external representations. This scheme utilizes a pre-trained full-precision network as a teacher with the same architecture as the BNN to be trained. This scheme distills each binarized convolutional layer, aiming to align the output representations between the binarized and full-precision models to reduce the impact of binarization on information of representation.}

\begin{algorithm}[b!]
    \caption{BNN Training Process of the proposed DIR-Net.} 
    \label{algo_forward}
    \begin{algorithmic}[1]
        \STATE \textbf{Require}: the input data $\mathbf a \in\mathbb R^{n}$, pre-activation $z \in\mathbb R$, full-precision weights $\mathbf w \in\mathbb R^{n}$.
        \STATE {\textbf{Forward propagation}}
        \STATE \quad Compute binary weight by IMB [Eq.~(\ref{eq:Libra-PB})]:\\
        \quad \quad $\hat{\mathbf w}_{\text{std}}=\frac{\mathbf w-\overline{\mathbf w}}{\sigma(\mathbf w-\frac{{\mathbf w}_{1}}{n}))}$,\quad $s=\operatorname{round}(\log_2 \frac{{||\hat{\mathbf w}_{\text{std}}||_{1}}}{n})$\\
        \quad \quad $\mathbf{Q_w}=\mathbf{B_{w}}\ll\gg s =\operatorname{sign}(\hat{\mathbf w}_{\text{std}})\ll\gg s$\\
        \STATE \quad Compute balanced binary input data [Eq.~(\ref{eq:Libra-PB})]:\\
        \quad \quad $\mathbf{Q_a}=\mathbf{B_{a}}=\operatorname{sign}(\mathbf a)$;
        \STATE \quad Calculate the output: ${z}=(\mathbf{B_{w}}\odot \mathbf{B_{a}})\ll\gg s$
        \STATE {\textbf{Back propagation}}
        \STATE \quad Update the $g'(\cdot)$ via DTE:\\
        \quad \quad Get current $t$ and $k$ by Eq.~(\ref{eq:DTE})\\
        \quad \quad Update the $g'(\cdot)$:\quad $g'(x)=kt(1-\tanh^2(tx))$
        \STATE \quad Calculate the gradients \wrt $\mathbf a$:\\
        \quad \quad $\frac{\partial{\mathcal{L}}}{\partial{\mathbf a}}=\frac{\partial{\mathcal{L}}}{\partial \mathbf{Q_a}}g'({\mathbf a})$\\
        \STATE \quad Calculate the gradients \wrt $\mathbf w$:\\
        \quad \quad $\frac{\partial{\mathcal{L}}}{\partial{\mathbf w}}=\frac{\partial{\mathcal{L}}}{\partial \mathbf{Q_w}}g'(\hat{\mathbf w}_{\text{std}})2^{s}$\\
        \STATE {\textbf{Parameters Update}}
        \STATE \quad {Update the network loss by RBD [Eq.~(\ref{eq:total_loss})]:}\\
        \quad \quad {Calculate the loss $\mathcal{L} = \mathcal{L}_{\text{CE}} + \gamma \mathcal{L}_{\text{RBD}}$}\\
        \quad \quad Update $\mathbf w$: $\mathbf w=\mathbf w-\eta\frac{\partial{\mathcal{L}}}{\partial{\mathbf w}}$, where $\eta$ is learning rate.
    \end{algorithmic}
\end{algorithm}

{Specifically, during the forward propagation, we save the output representation of the binarized convolutional layer in the BNN and that of the corresponding teacher model, respectively.
And inspired by \cite{martinez2020training}, we use the attention form in the distillation process for representations of the teacher and student networks, which mainly pays attention to the inner relation of representations and eliminates the influence of scale.
When we substitute it into Eq.~(\ref{eq:output_error}), the error for the $\ell$-th layer can be expressed as:}
\begin{equation}
\label{eq:output_attention}
\mathcal{E}_{{\textbf{z}^2}}^\ell=\frac{{\textbf{z}^\ell}^2}{\|{\textbf{z}^\ell}^2\|_{2}}-\frac{{\textbf{z}_\text{fp}^\ell}^2}{\|{\textbf{z}_\text{fp}^\ell}^2\|_{2}},
\end{equation}
{then we take the L2 normalization of the representation error $\mathcal{E}^\ell_{{\textbf{z}}^2}$ as the distillation loss term of the layer and sum the loss terms:}
\begin{equation}
\label{eq:loss-r2b}
\mathcal{L}_\text{RBD}
= \sum^{L}_{\ell=1} \left\|\mathcal{E}^\ell_{\textbf{z}^2}\right\|_2
= \sum^{L}_{\ell=1} \left\| \frac{\mathbf{z}^{\ell\ 2}}{\|\mathbf{z}^{\ell\ 2}\|_2} - \frac{\mathbf{z}^{\ell\ 2}_\text{fp}}{\| \mathbf{z}^{\ell\ 2}_\text{fp}\|_2} \right\|_2.
\end{equation}
{The losses from optimization techniques are summed to update the weight jointly during the backward propagation, which can be expressed as: }
\begin{equation}
\label{eq:total_loss}
\mathcal{L} = \mathcal{L}_{\text{CE}} + \gamma \mathcal{L}_{\text{RBD}} ,
\end{equation}
{where $\mathcal{L}_{\text{CE}}$ denotes the Cross-Entropy (CE) loss of the BNN and $\gamma$ is a hyperparameter to control distillation impact, set to 0.1 as default.}

{Our RBD scheme introduces an additional full-precision teacher model, which can be regarded as the most ideal information-retaining version of the BNN to be trained. Then, representation alignment is performed by distilling the constructed error metric in the form of attention.}

\subsection{Analysis and Discussions}

The training process of our DIR-Net is summarized in Algorithm~\ref{algo_forward}. In this section, we will analyze DIR-Net from different aspects.

\subsubsection{Complexity Analysis}
{Since IMB and DTE are applied during the training process, there is no extra operation for binarizing activations in DIR-Net.} And in IMB, with the novel shift-based scalars, the computation costs are reduced compared with the existing solutions with 32-bit scalars (\eg, XNOR-Net, and LQ-Net), as shown in Table~\ref{tab:flops}. Moreover, we further test the real speed of deployment on hardware and we showcase the results in Sec.~\ref{deploy_efficiency}.

\begin{table}[h!]
    \caption{The additional 32-bit operations consumed by different binarization methods.}
    \vspace{-0.1in}
    \centering
    \setlength{\tabcolsep}{8.5mm}{
    \begin{threeparttable}
    \begin{tabular}{lcc}
        \toprule
        Method & Float Operations & \tabincell{c}{Bitwise Operations}\\
        \midrule
        XNOR-Net& $C_{\text{1}}$&$C_{\text{1}}\times C_{\text{2}}$\\ 
        LQ-Net&$C_{\text{1}}$&$C_{\text{1}}\times C_{\text{2}}$\\ 
        Ours&0&$C_{\text{1}}\times C_{\text{2}}+C_{\text{1}}$\\
        \bottomrule
    \end{tabular}
    \begin{tablenotes}
    \footnotesize
    \item[*] $C_{\text{1}}=w_{\text{out}}\times h_{\text{out}}\times c_{\text{out}}$ and  $C_{\text{2}}=w_{\text{k}}\times h_{\text{k}}\times c_{\text{in}}$, where $c_{\text{out}} $, $ c_{\text{in}} $, $ w_{\text{k}} $, $ h_{\text{k}} $, $ w_{\text{out}} $, $ h_{\text{out}} $ denote the number of output channels, input channels, kernel width, kernel height, output width, and output height, respectively. The Bitwise operation mainly consists of XNOR, Bitcount and Shift.
    \end{tablenotes}
    \end{threeparttable}
    }
    \label{tab:flops}
\end{table}

\subsubsection{Stabilize Training}
\begin{wrapfigure}[12]{r}{0.4\textwidth}
\vspace{-1.in}
\includegraphics[width=1.0\textwidth]{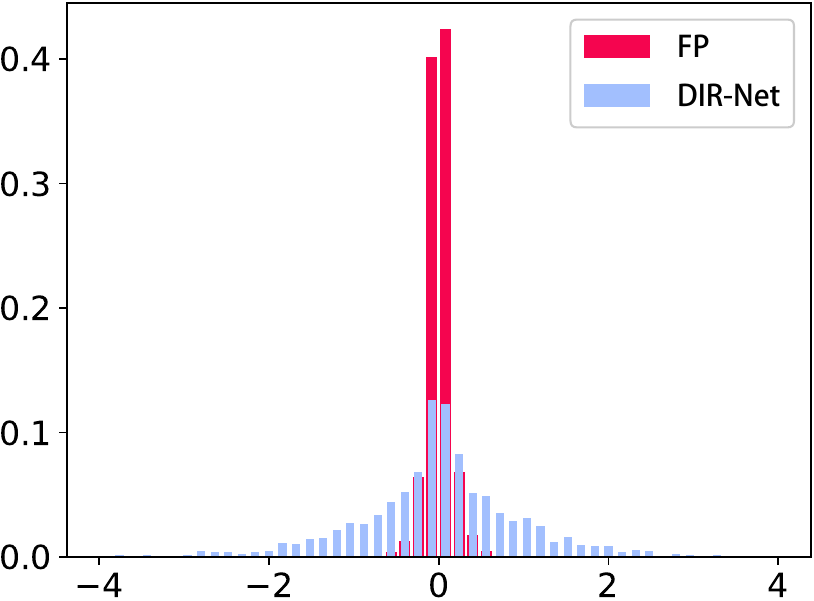}
\vspace{-0.2in}
\caption{Full-precision weights (in red) in neural networks have a small data range and always gather around 0, and thus their signs are highly possible to flip in the backward propagation. The DIR-Net balances and standardizes the weights (in blue) before the binarization for stabilizing training.}
\label{distribution}
\end{wrapfigure}
In IMB, weight standardization is introduced for stabilizing training, which avoids fierce changes of binarized weights.
Fig.~\ref{distribution} shows the data distribution of weights without standardization, obviously more concentrated around 0. This phenomenon means the signs of most weights are easy to change during the process of optimization, which directly causes unstable training of binary neural networks. By redistributing the data, weight standardization implicitly sets up a bridge between the forward IMB and backward DTE, contributing to a more stable training of binary neural networks. 
Moreover, the proposed DTE also stabilizes the training by not only ensuring the updating capability of networks, but also preventing the estimator from being too steep, and thus avoid the gradients from being excessively enlarged.

\section{Experiments}
\label{sec:Exp}

We perform image classification on two benchmark datasets: CIFAR-10~\cite{CIFAR} and ImageNet (ILSVRC12)~\cite{Deng2009ImageNet} and object detection task on PASCAL VOC~\cite{Everingham10} and COCO~\cite{lin2014microsoft} datasets to evaluate our DIR-Net and compare it with other recent SOTA methods over various architectures.

\textbf{DIR-Net:} We implement our DIR-Net based on PyTorch since it has a high degree of flexibility and a powerful automatic differentiation mechanism. To build a binarized model, we just use the binary convolutional layers binarized by our method instead of the convolutional layers of the original models.

\textbf{Network Structures:} We evaluate our DIR-Net performance on mainstream and compact CNNs structures, including VGG-Small~\cite{LQ-Net}, ResNet-18, ResNet-20 on CIFAR-10, and ResNet-18, ResNet-34~\cite{he2016deep}, MobileNetV1~\cite{mobilenet}, EfficientNet-B0~\cite{tan2019efficientnet}, and DARTS~\cite{liu2018darts} on ImageNet dataset in our experiments.
{To verify the versatility of our method, we also evaluate our DIR-Net on networks with normal structure and ReActNet~\cite{liu2020reactnet} structure, the latter includes several structure designs and is specifically proposed for binarization. We guarantee the architecture of DIR-Net in the reported results is consistent with existing SOTA competitors.}
We binarize all convolutional and fully connected layers except the first and last one, and keep the 1x1 convolution to full-precision in EfficientNet and DARTS. 

\textbf{Hyper-parameters and other setups:} 
In order to evaluate our DIR-Net on different CNN structures, we mostly apply the original hyper-parameter settings and training steps in their papers, and follow exactly the presented results in their paper~\cite{xnornet,LQ-Net,Liu_2018_ECCV,IRNet,liu2020reactnet,wang2020bidet}. 
Specifically, for experiments on CIFAR-10, we train the models for up to 400 epochs. The learning rate starts at 1e-1 and decays to 0 during training by the Cosine Annealing scheduler~\cite{loshchilov2016sgdr}. 
For experiments on ImageNet, we train the models for up to 250 epochs. The learning rate starts at 1e-2 and decays to 0 during training by the Cosine Annealing scheduler.
The weight decay of 1e-4 and batch size of 128 are adopted following the original paper, and SGD is applied as the optimizer with a momentum of 0.9. 
{For experiments on Pascal VOC and COCO datasets, we train the models for up to 300000 iterations. The learning rate starts at 1e-3 and decays at 30000th, 80000th, 200000th iteration, the weight decay of 0 and batch size of 32 are adopted, and Adam is applied as the optimizer.}  
{And we evaluate the results of DIR-Net in all comparative experiments with 3 different random seeds and report their mean values, and discuss the standard deviations of results in the corresponding specific sections.}

\subsection{Ablation Study}
In this section, we evaluate the performance and effects of our proposed IMB and DTE on BNN.

\subsubsection{Effect of IMB}
Our proposed IMB adjusts the distribution of weights to maximize the information entropy of binary weights and binary activations in the network. Due to the balance operation before binarization, the binary weight parameters of each layer in the DIR-Net have the maximal information entropy. As for binary activations affected by binary weights in DIR-Nets, the maximization of its information entropy is also guaranteed.

\begin{figure}[tp!]
\subfigure[]{
\centering
\includegraphics[width=0.46\textwidth]{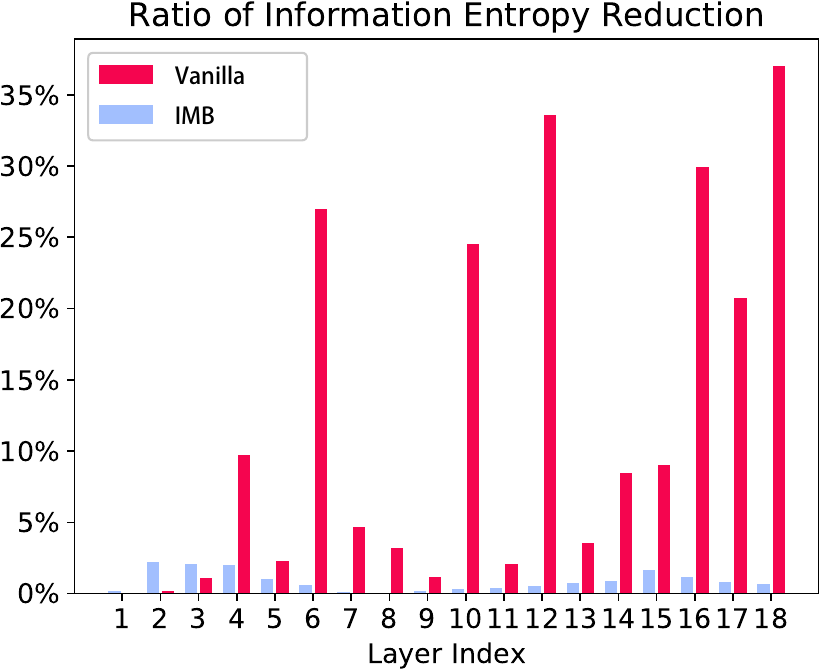}
\label{fig:activation_entropy}}
\vspace{0.1in}
\subfigure[]{
\centering
\includegraphics[width=0.4\textwidth]{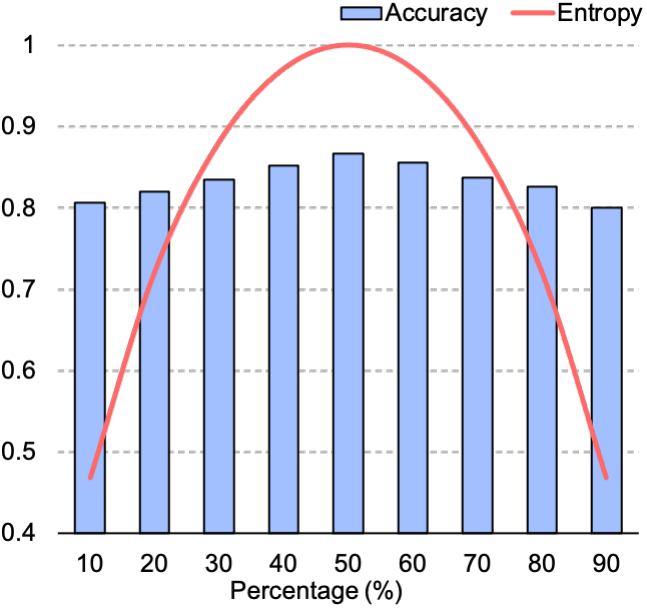}
\label{fig:ent_acc}}
\vspace{-0.25in}
\caption{(a) IMB's effect on information entropy of activations in each layer of ResNet-20. 
(b) The effect of activation information entropy on accuracy for ResNet-20 on CIFAR-10 dataset, where the $x$-axis coordinate indicates the percentage of activations that less than 0.}
\end{figure}
In order to illustrate the information retention capability of IMB, in Fig.~\ref{fig:activation_entropy}, we show the information entropy reduction of each layer's binary activations in the network quantized by IMB and vanilla binary neural network respectively. 
As the figure shown, vanilla binarization results in a great decrease in the information entropy of binary activations. It is notable that the information loss seems to accumulate across layers in the forward propagation.
Fortunately, in the IMB quantized networks, the information entropy of activations of each layer is close to the maximal information entropy under the Bernoulli distribution. IMB can achieve information retention of the binary activations in each layer.

We further evaluate the impact of information on our DIR-Net in detail. The information in DIR-Net is defined by Eq.~(\ref{ie}), and can be adjusted by changing the mean value of activations. 
Fig.~\ref{fig:ent_acc} presents the relationship between the information of weights in DIR-Net and the final accuracy. The information entropy of binarized activations is determined by the percentage of full-precision counterparts that less than 0.
The results show that information entropy is almost positively correlated with network accuracy. And when the information is maximized, the BNN achieves the highest accuracy, which verifies the effectiveness of IMB.
Therefore, information entropy is an important indicator to measure the amount of information that BNN holds, and we can improve BNNs' performance by maximizing the entropy.

In addition, we analyze the impact of IMB on binarization errors. In Table~\ref{ablation_exp_error}, we compare binarized networks that apply different approaches to quantize weights (activations are binarized directly for the sake of fairness), including vanilla binarization, XNOR binarization, and our IMB. 
The XNOR binarization uses 32-bit scalars while our IMB uses integer shift-based scalars.
Compared with vanilla binarization, BNNs quantized by XNOR and IMB have a much smaller binarization error since the usage of scalars.
Our IMB further eliminates all floating-point scalars (1.3e3) and related floating-point computation compared with XNOR binarization, while the binarization error only increases by 5\% (0.1e4) but the quantized network enjoys better accuracy (84.9\%).
The results show that our IMB has a better balance between inference speedup and binarization error minimization.

\begin{table}[bht]
    \caption{Ablation study for binarization error on ResNet-20.}
    \vspace{-0.1in}
    \label{ablation_exp_error}
    \centering
    \setlength{\tabcolsep}{2.5mm}
    {\begin{tabular}{lcccc}
        \toprule
        Method&\tabincell{c}{Bit-width (W/A)}&\tabincell{c}{Error}&\tabincell{c}{Float Scalar}&\tabincell{c}{Accuracy (\%)}\\
        \midrule
        Full-Precision &32/32& - & - &91.7\\
        Binary&1/1&5.1e4& 0&83.8\\
        XNOR&1/1&\textbf{1.9e4}&1.3e3 &84.8\\
        IMB (Ours)&{1/1}&2.0e4& \textbf{0}&\textbf{84.9}\\
        \bottomrule
    \end{tabular}
    }
    \vspace{-0.1in}
\end{table}

\subsubsection{Effect of DTE}
Firstly, we discuss the setting of the parameter $t_\epsilon$ in the DTE, which determines the degree of updating capability that DTE maintains. As mentioned in section~\ref{sec:dte}, we control the value of the parameter $t_\epsilon$ to ensure that at least $\epsilon$ of parameters are updated during the whole training process. However, if the value of $\epsilon$ is set too large, the gradients will be not accurate since the gap between the estimator and the $\operatorname{sign}$ function is huge. 
Table~\ref{epsilon_exp} shows the accuracy of DIR-Net under different $t_{\epsilon}$ settings, based on the ResNet-20 architecture and the CIFAR-10 dataset. The results show that the accuracy increases with the decreasing of the value of $\epsilon$ in a considerable range (approximately 100\%-10\%).
And the models which properly ensure the updating capability perform better than that do not control the lower limit of updating capability at all ($\epsilon$ is set to 0\%). 
The results show that compared with the estimator that simply minimizes the gradient error, appropriately improving the minimum updating capability of binarized model and keeping enough parameters updating during training are more helpful to improve BNN performance.
Therefore, in our experiment, we empirically set the value of $\epsilon$ to 10\% to achieve a good trade-off between accurate gradient and updating capability.

\begin{table}[bht]
    \caption{{The effect of parameter $\epsilon$ on accuracy.}}
   \vspace{-0.1in}
    \label{epsilon_exp}
    \centering
    \setlength{\tabcolsep}{1.mm}{
    \begin{tabular}{lccccccccccc}
        \toprule
        {\textbf{$\epsilon$} \textbf{(\%)}}&{100}&{90}&{80}&{70}&{60}&{50}&{40}&{30}&{20}&{10}&{0}\\
        \midrule
        \tabincell{c}{{\textbf{Accuracy (\%)}}}& {84.2}&  {84.4}&  {84.7}&  {84.6}&  {85.1}&  {85.7}&  {85.9}&  {86.2}&  {86.7} &  {\textbf{86.8}}&  {86.5}\\
        \bottomrule
    \end{tabular}
    }
    \vspace{-0.1in}
\end{table}

\begin{wrapfigure}[15]{r}{0.6\textwidth}
\includegraphics[width=1.0\textwidth]{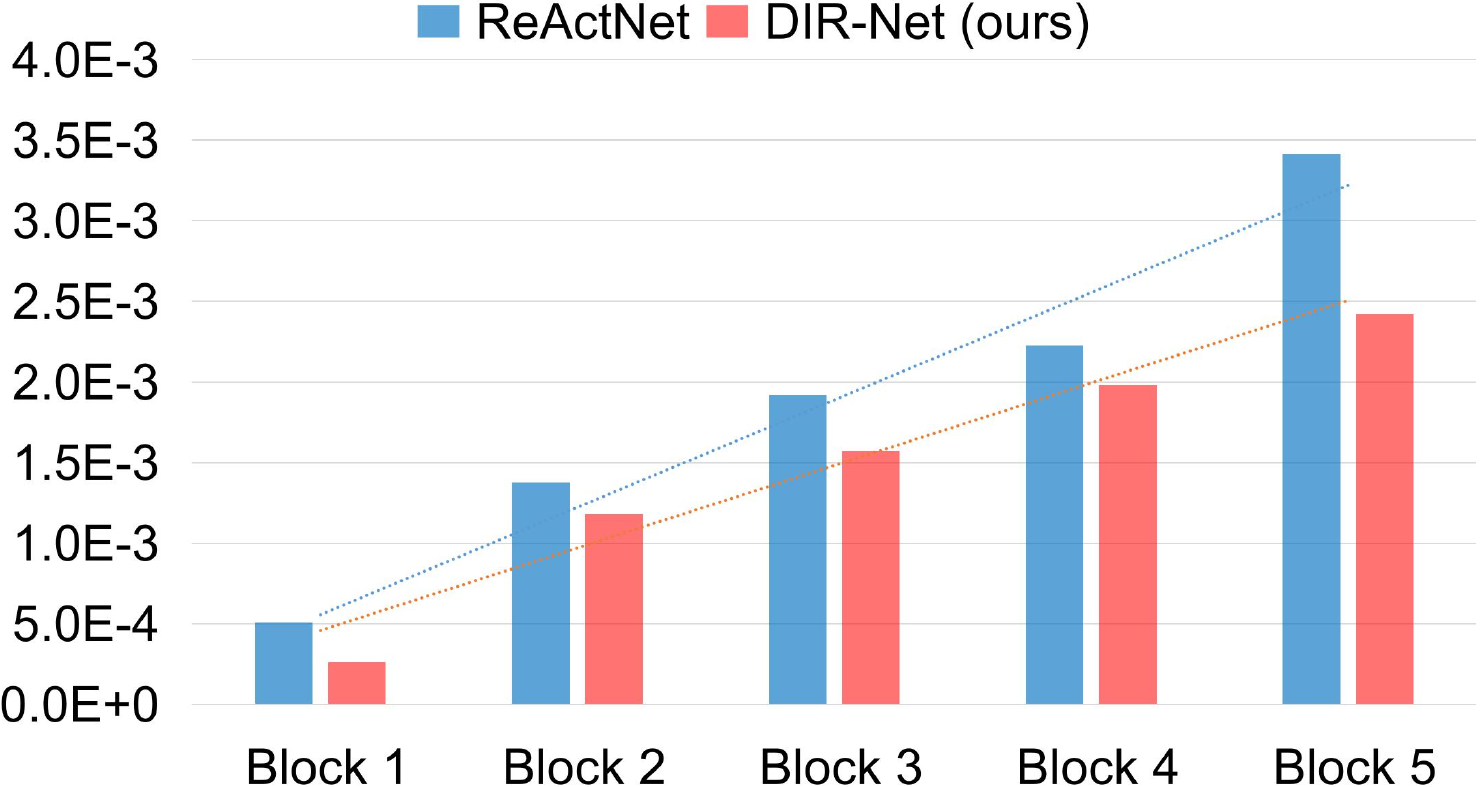}
\vspace{-0.25in}
\caption{Comparison of representation errors of ResNet-18 quantized by ReActNet and DIR-Net. We present the representation errors of the last convolutional layer in residual blocks to compare with a full-precision teacher with the same structure.}
\label{fig:error}
\end{wrapfigure}
\begin{figure}[th]
\begin{center}
\includegraphics[width=0.8\linewidth]{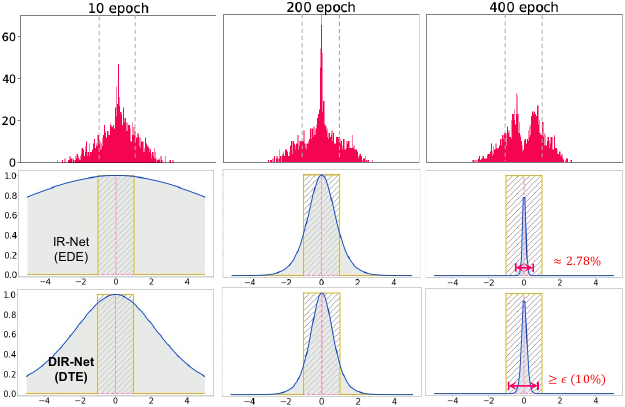}
\end{center}
\vspace{-0.15in}
\caption{The proposed distribution-sensitive DTE takes full account of the impact of weight distributions (after standardization) in different epochs (10, 200 and 400) during training. The weight distributions are shown in the upper part, the middle and bottom parts are the comparisons of estimators in DIR-Net (DTE) and existing IR-Net (EDE).
Taking DTE as an example, the blue line is the derivative of DTE, the yellow and the pink lines are the derivative of the STE and $\operatorname{sign}$ function, respectively. And the shade indicates the error between the derivative of DTE/STE and that of the $\operatorname{sign}$ function.
Compared with existing EDE, DTE further ensures that at least $\epsilon (10\%)$ weights can always keep updating capability during training. }
\label{fig:DTE_w_d} 
\end{figure}

Then, in order to illustrate the effect and necessity of DTE, we show the distribution of weight parameters in different epochs of training in Fig.~\ref{fig:DTE_w_d}. The figures in the first row show the distribution of data, and the ones in the second and third rows show the curve of the corresponding derivative of existing EDE and our DTE, respectively. Among the derivative curves, the blue curves represent the derivative of EDE or DTE and the yellow ones represent the derivative of STE (with clipping). Obviously, in the first stage (epoch 10 to epoch 200 in Fig.~\ref{fig:DTE_w_d}) of DTE, there are lots of data beyond the range of $[-1, +1]$, thereby the estimator should have a larger effective updating range to ensure the updating capability of the BNN. In addition, the peakedness of weight distribution is high and a large amount of data is clustered near zero when training begins. DTE keeps the derivative close to the $\operatorname{identity}$ function at this stage to avoid the derivative value near zero being too large, thereby preventing severe unstable training. Fortunately, as binarization is introduced into training, the weights will be gradually redistributed around $-1/+1$ in the later stages of training. Therefore, we can slowly increase the derivative value and approximate the standard $\operatorname{sign}$ function to avoid gradient mismatch.
The visualized results show that our DTE approximation for backward propagation is consistent with the real data distribution, which is critical to improving the accuracy of networks. 
Moreover, compared to the existing EDE, the improvement of DTE lies in the ability to maintain the network's updating capability throughout the training process, especially in the later stages of training. As shown in Fig.~\ref{fig:DTE_w_d}, at the 400 epoch, the derivative of EDE is almost the same as the $\operatorname{sign}$ function (the pink line) and only $2.78\%$ of weights can be updated, while the proposed DTE ensures at least $\epsilon$ (default set as 10\%) of weights can be continuously updated.

\subsubsection{Effect of RBD}
{The proposed RBD aims to align the representation between the full-precision and the binarized ones by distilling the corresponding activation outputs with an especially designed loss function. By aligning the inner relations, the impact of high discrete representation is further reduced with more information retained in the network.}

{Fig.~\ref{fig:error} shows the phenomenon of error accumulation in ReActNet and DIR-Net. It shows the average representation error of 30 random samples after every four layers, which are calculated by Eq.~(\ref{eq:output_error}) and takes the absolute sum of each convolutional layer. 
The binarized network always loses information during propagation, while the corresponding full-precision network is not affected by binarization and is considered to be well-trained. Thus, we induce the full-precision representations as external knowledge to better reduce the difficulty of training binarized networks. We can see from the figure that based on ResNet-18, for ReActNet, the representation error between binarized parameters and corresponding full-precision parameters begins with 5.1e-4 after the first convolution layer, and increases gradually to 3.4e-3. But our DIR-Net is equipped with RBD, and the distillation effectively closes the distance to the full-precision counterpart which is 29.1\% lower than ReActNet after the last block. And thus stabilizes the training process of binarized networks to get higher results.}

\subsubsection{Ablation Performance}
We further evaluate the performance of different parts of DIR-Net using the ResNet-20 architecture on the CIFAR-10 dataset, which helps understand how DIR-Net works in practice. Table~\ref{ablation_exp} presents the accuracy of networks with different settings. 
As the Table~\ref{ablation_exp} shown, both IMB and DTE can improve the accuracy.
For the IMB, both the the standardization of weight data in IMB plays an important role.
The accuracy of BNN with weight standardization and bit-shift scalars is 0.5\% and 0.8\% higher than that of naive BNN, respectively, while the IMB version achieves 1.1\% gain.
For the DTE, the key to improving BNN performance is the cooperation of its two stages.
The performance of BNN that only applies the stage1 of DTE is even not as good as the naive BNN with $\operatorname{clip}$ approximation.
And the results of BNNs only applying stage2 show that, the estimator that ensures a certain updating capability always be retained ($\epsilon=10\%$) is more effective to BNN compared with the estimator that does not control the lower limit ($\epsilon=0\%$). The phenomenon proves the motivation of DTE, which is specifically ensuring the minimum update capability of the estimator during the training process.
The usage of DTE takes 1.7\% gain to BNN.
{For the RBD, in Table~\ref{ablation_exp}, we show the results with or without RBD, using different forms of loss functions, including Mean Squared Error (MSE) loss and KL divergence loss~\cite{hinton2015distilling}. See from the results, RBD with Eq.~(\ref{eq:output_attention}) gets the most obvious improvements which boost the performance from 83.8\% to 86.2\%. Meanwhile, distilling the full-precision representation with MSE and KL divergence loss functions can also help with the training and the final results (1.1\% and 1.3\% improvements, respectively), which confirms that inducing the external knowledge of full-precision and aligning the representations can reduce the negative impact of error accumulation caused by highly discrete parameters.
Moreover, the improvements in IMB, DTE, and RBD can be superimposed, hence we can train binary neural networks with high accuracy using our method.}

\begin{table}[th]
    \caption{Ablation study for DIR-Net.}
  \vspace{-0.1in}
    \label{ablation_exp}
    \centering
    \setlength{\tabcolsep}{4.5mm}{
    \begin{tabular}{lcc}
        \toprule
        Method&\tabincell{c}{Bit-width (W/A)}&Acc. (\%)\\
        \midrule
        {Full-Precision}&{32/32}&{91.7}\\
        Binary&1/1&83.8\\
        IMB (w/o weight standardization)&1/1&84.3\\
        IMB (w/o shift-based scalars)&1/1&84.6\\
        IMB&1/1&84.9\\
        DTE (stage1)&1/1&83.6\\
        DTE (stage2, $\epsilon=0\%$)&1/1&84.9\\
        DTE (stage2, $\epsilon=10\%$)&1/1&85.1\\
        DTE&1/1&85.5\\
        {RBD (MSE)} & {1/1} & {84.9} \\
        {RBD (KL divergence)} & {1/1} & {85.1} \\
        {RBD}&{1/1}&{86.2} \\
        {DIR-Net (IMB \& DTE \& RBD)}&{1/1}&{\textbf{89.0}}\\
        \bottomrule
    \end{tabular}
    }
    \vspace{-0.2in}
\end{table}

\subsection{Comparison with SOTA methods}

We have performed a complete evaluation of the DIR-Net by comparing it with the existing SOTA methods.

\begin{table}[t]
\caption{Performance comparison on CIFAR-10.}
  \vspace{-0.1in}
    \label{cifar}
    \centering
    \begin{threeparttable}
    \setlength{\tabcolsep}{4.5mm}{
    \begin{tabular}{llcc}
        \toprule
        Topology&Method&\tabincell{c}{Bit-width (W/A)}&\tabincell{c}{Acc. (\%)}\\
        \midrule
        \multirow{7}{*}{ResNet-18}&Full-Precision&32/32&93.0\\
        &{Bi-Real}&{1/1}&{89.1}\\
        &{XNOR}&{1/1}&{90.2}\\
        &RAD&1/1&90.5\\
        &IR-Net&1/1&{91.5}\\
        &{ReActNet}&{1/1}&{92.3}\\
        &{DIR-Net (ours)}&{1/1}&{\textbf{92.8}$_{\pm0.14}$}\\
        \midrule
        \multirow{13}{*}{ResNet-20}&Full-Precision&32/32&91.7\\
        &DoReFa&1/1&79.3\\
        &DSQ&1/1&84.1\\
        &{Bi-Real}&{1/1}&{85.7}\\
        &{IR-Net}&{1/1}&{86.5}\\
        &{ReActNet}&{1/1}&{87.9}\\
        &{DIR-Net (ours)}&{1/1}&{\textbf{89.0}$_{\pm0.07}$}\\
        \cmidrule(r){2-4}
        &Full-Precision&32/32&91.7\\
        &DoReFa&1/32&90.0\\
        &LQ-Net&1/32&90.1\\
        &DSQ&1/32&90.2\\
        &{IR-Net}&{1/32}&{90.8}\\
        &{DIR-Net (ours)}&{1/32}&{\textbf{91.3}$_{\pm0.06}$}\\
        \midrule
        \multirow{7}{*}{VGG-Small}&Full-Precision&32/32&91.7\\
        &LAB&1/1&87.7\\
        &XNOR&1/1&89.8\\
        &BNN&1/1&89.9\\
        &RAD&1/1&90.0\\
        &{IR-Net}&{1/1}&{90.4}\\
        &{DIR-Net (ours)}&{1/1}&{\textbf{91.1}$_{\pm0.10}$}\\
        \bottomrule
    \end{tabular}}
    \vspace{-0.1in}
    \end{threeparttable}
\end{table} 

\subsubsection{Image Classification Tasks}

\textbf{CIFAR-10 dataset}. Table~\ref{cifar} lists the performance of different methods on the CIFAR-10 dataset, and we compare our DIR-Net with these methods on various widely used architectures, such as ResNet-18~\cite{ResNet-18-project}, ResNet-20~\cite{ResNet-20-project}, and VGG-Small.
{We show the comparison with results of RAD~\cite{Regularize-act-distribution}, IR-Net~\cite{IRNet}, and ReActNet~\cite{liu2020reactnet} over ResNet-18, DoReFa-Net~\cite{dorefa}, LQ-Net~\cite{LQ-Net}, DSQ~\cite{DSQ}, IR-Net~\cite{IRNet}, and ReActNet~\cite{liu2020reactnet} over ResNet-20,} BNN~\cite{hubara2016binarized}, LAB~\cite{Loss-Aware-BNN}, RAD~\cite{Regularize-act-distribution}, XNOR-Net~\cite{xnornet} and IR-Net~\cite{IRNet} over VGG-Small.

In all cases in the table, our proposed DIR-Net has the highest accuracy. Moreover, in the case of using the ResNet architecture, our DIR-Net has a significant improvement compared to the existing SOTA methods when using 1-bit weights and 1-bit activations (1W/1A). For example, with the 1W/1A bit-width setting, {the accuracy of our method is improved by 4.9\% compared to DSQ on ResNet-20, and the gap between the full-precision counterpart is reduced to 2.7\%.} Compared with the IR-Net, our DIR-Net performs better since it further ensures enough parameters can be updated during the training process, {and it can outperform IR-Net by generally 1.2\% with different backbones on 1W/1A bit-width.}
{Moreover, among the results of all network architectures, the standard deviation of the results using different random numbers is less than 0.14\%, and it is even as low as 0.10\% on the VGG-Small, which is much lower than the improvement against existing SOTA methods on these architectures (at least 0.5\%).}
The results show that the improvement of our DIR-Net is robust and can stably improve the network performance under various settings.

\noindent\textbf{ImageNet dataset}. We study the performance of DIR-Net over ResNet-18, ResNet-34, MobileNetV1, DARTS, and EfficientNet-B0 structures on the large-scale ImageNet dataset. 
Table~\ref{imagenet} lists the comparison with several SOTA quantization methods, including BWN~\cite{xnornet}, HWGQ~\cite{DBLP:journals/corr/abs-1708-08687}, TWN~\cite{DBLP:journals/corr/LiL16}, LQ-Net~\cite{LQ-Net}, DoReFa-Net~\cite{dorefa}, ABC-Net~\cite{ABCNet}, Bi-Real~\cite{Liu_2018_ECCV}, XNOR++~\cite{XNOR++}, BWHN~\cite{DBLP:journals/corr/abs-1802-02733}, SQ-BWN and SQ-TWN~\cite{Dong2017Learning}, 
PCNN~\cite{DBLP:journals/corr/abs-1811-12755}, BONN~\cite{gu2019bayesian}, Si-BNN~\cite{wang2020sparsity}, Real-to-Bin~\cite{martinez2020training}, MeliusNet~\cite{bethge2021meliusnet}, and ReActNet~\cite{liu2020reactnet}.

As shown in Table~\ref{imagenet}, when only quantizing weights over ResNet-18 with 1-bit weights, DIR-Net greatly exceeds most other methods, {and even outperforms the TWN with 2-bit weights by a notable 5.7\%. 
Meanwhile, DIR-Net outperforms IR-Net 0.7\% on Top-1 accuracy and 0.9\% on Top-5 accuracy based on ResNet-34 architecture using 1W/32A setting.} 
Moreover, while using the 1W/1A setting, our DIR-Net also surpasses the SOTA binarization methods. {The Top-1 accuracy of our DIR-Net is apparently higher than that of the ReActNet (65.9\% vs. 66.5\% for ResNet-18) and Si-BNN (63.3\% vs. 67.9\% for ResNet-34).} The results prove that our DIR-Net is more competitive than the existing binarization methods.

\begin{table}[htp]
    \caption{Performance comparison on ImageNet.}
    \label{imagenet}
    \vspace{-0.1in}
    \centering
    \begin{threeparttable}
    {
    \setlength{\tabcolsep}{2.7mm}{
    \begin{tabular}{llccc}
        \toprule
        Topology&Method&\tabincell{c}{Bit-width (W/A)}&Top-1(\%)&Top-5(\%)\\
        \midrule
        \multirow{27}{*}{ResNet-18}&Full-Precision&32/32&69.6&89.2\\
        &ABC-Net&1/1&42.7&67.6\\
        &XNOR&1/1&51.2&73.2\\ 
        &BNN+&1/1&53.0&72.6\\
        &DoReFa&1/2&53.4&--\\
        &Bi-Real&1/1&56.4&79.5\\
        &XNOR++&1/1&57.1&79.9\\
        &PCNN&1/1&57.3&80.0\\
        &{IR-Net}&{1/1}&{58.1}&{80.0}\\
        &BONN&{1/1}&{58.3}&{81.6}\\
        &Si-BNN&{1/1}&{59.7}&{81.8}\\
        &{Real-to-Bin}&{1/1}&{65.4}&{86.2}\\
        &{ReActNet}&{1/1}&{65.9}&{--}\\
        &{DIR-Net\footnotemark[1] (ours)}&{1/1}&{60.4}&{81.9}\\
        &{DIR-Net\footnotemark[2] (ours)}&{1/1}&{\textbf{66.5}$_{\pm0.10}$}&{\textbf{87.1}}\\
        \cmidrule(r){2-5}
        &Full-Precision&32/32&69.6&89.2\\
        &SQ-BWN&1/32&58.4&81.6\\
        &BWN&1/32&60.8&83.0\\
        &HWGQ&1/32&61.3&83.2\\
        &TWN&2/32&61.8&84.2\\
        &SQ-TWN&2/32&63.8&85.7\\
        &BWHN&1/32&64.3&85.9\\
        &{IR-Net}&{1/32}&{66.5}&{86.8}\\
        &{DIR-Net (ours)}&{1/32}&{\textbf{67.5}$_{\pm0.06}$}&{\textbf{87.9}}\\
        \midrule
        \multirow{14}{*}{ResNet-34}&Full-Precision&32/32&73.3&91.3\\
        &ABC-Net&1/1&52.4&76.5\\
        &Bi-Real&1/1&62.2&83.9\\
        &{IR-Net}&{1/1}&{62.9}&{84.1}\\
        &{Si-BNN}&1/1&63.3&84.4\\
        &{ReActNet}&{1/1}&{67.3}&{87.9}\\
        &{DIR-Net\footnotemark[1] (ours)}&{1/1}&{64.1}&{85.3}\\
        &{DIR-Net\footnotemark[2] (ours)}&{1/1}&{\textbf{67.9}$_{\pm0.09}$}&{\textbf{88.2}}\\
        \cmidrule(r){2-5}
        &Full-Precision&32/32&73.3&91.3\\
        &{ABC-Net}&{1/32}&{68.8}&{86.1}\\
        &{Bi-Real}&{1/32}&{69.7}&{88.9}\\
        &{Si-BNN}&{1/32}&{70.1}&{89.7}\\
        &IR-Net&1/32&70.4&89.5\\
        &{DIR-Net (ours)}&{1/32}&{\textbf{71.1}$_{\pm0.03}$}&{\textbf{90.4}}\\
        \midrule
        \multirow{7}{*}{{DARTS}}&{Full-Precision}&32/32& 73.3& 91.3\\
        &{BNN} & {1/1} & {52.2} & {76.6} \\
        &{Bi-Real} & {1/1} & {61.5} & {83.8} \\
        &{IR-Net}&{1/1}& {62.1}& {84.2}\\
        &{ReActNet}&{1/1}&{65.1}&{86.4}\\
        &{DIR-Net\footnotemark[1] (ours)}&{1/1}& {63.3}& {85.1}\\
        &{DIR-Net\footnotemark[2] (ours)}&{1/1}& {\textbf{65.6}$_{\pm0.12}$}&{\textbf{87.2}}\\
        \midrule
        \multirow{7}{*}{{EfficientNet}}&{Full-Precision}&32/32& 76.2& 92.7\\
        & {BNN} & {1/1} & {52.7} & {76.5} \\
        & {Bi-Real} & {1/1} & {58.7} & {81.3} \\
        &{IR-Net}&{1/1}&{60.1}&{82.6}\\
        &{ReActNet}&{1/1}&{63.5}&{85.1}\\
        & {DIR-Net\footnotemark[1] (ours)} & {1/1} & {63.0} & {84.8} \\
        &{DIR-Net\footnotemark[2] (ours)}&{1/1}& {\textbf{64.8}$_{\pm0.09}$}&{\textbf{86.2}}\\
        \midrule
        \multirow{7}{*}{{MobileNet}}&{Full-Precision}&32/32& 72.4&-- \\
        &{BNN}&1/1& 60.9& --\\
        &{MeliusNet22}&{1/1}&{63.6}&{84.7}\\
        &{MeliusNet29}&{1/1}&{65.8}&{86.2}\\
        &{MeliusNet42}&{1/1}&{69.2}&{88.3}\\
        &{ReActNet}&1/1& 69.5&-- \\
        &{DIR-Net (ours)}&{1/1}& {\textbf{70.6}$_{\pm0.16}$}& {\textbf{89.7}}\\
        \bottomrule
    \end{tabular}}
    }
    \begin{tablenotes}
    \footnotesize
    \item[1] Results of networks with normal structure.
    \item[2] Results of networks with ReActNet structure~\cite{liu2020reactnet}.
    \end{tablenotes}
    \end{threeparttable}
\end{table}

We further implemented our DIR-Net on more compact CNN structures, including DARTS, EfficientNet and MobileNet, and compared with other SOTA binarization methods.
Results in Table~\ref{imagenet} shows that our DIR-Net outperforms both vanilla BNN and Bi-Real Net over the DARTS and EfficientNet-B0 structures without any additional computational overheads and training steps.
{Over the DARTS structure, our DIR-Net surpasses Bi-Real by 2.8\% on Top-1 and 3.6\% on Top-5 accuracy, respectively, and surpasses ReActNet by 0.5\% and 0.8\%.
Over the EfficientNet-B0 structure, DIR-Net surpasses Bi-Real by 3.9\% Top-1 and 4.2\% Top-5 accuracy, and surpasses ReActNet by 0.9\% and 0.7\%. In both cases DIR-Net outperforms the basic BNN method by an convincing margin.}
As for the MobileNet structure, DIR-Net also performs well and surpasses the SOTA methods.
{Under the setting of 1W/1A, our DIR-Net only loses 1.8\% of the Top-1 accuracy compared with the full-precision counterpart, which is much better than other binarization methods.}
Experiments on these compact networks show that our binarization scheme is versatile and competitive in various structures.
{We also recorded the fluctuation of the DIR-Net Top-1 accuracy with different random seeds, where the maximum and minimum values of the standard deviation under various settings are 0.16\% and 0.03\%, respectively, and all the results stably exceeded the existing binarization methods.}

\begin{table}[htp]
    \caption{Performance comparison on PASCAL VOC.}
    \label{pascal-voc}
    \vspace{-0.1in}
    \centering
    {
    \setlength{\tabcolsep}{1.5mm}{
    \begin{tabular}{lcccc}
        \toprule
        {Framework (Backbone)} & {Input} & {Method} & {Bit-width (W/A)} & {mAP (\%)}\\
        \midrule
        \multirow{5}{*}{{SSD300 (VGG-16)}} & \multirow{5}{*}{{300$\times$300}} & {Full-Precision} & {32/32} & {72.4}\\
        & & {BNN} & {1/1} & {42.0}\\
        & & {XNOR} & {1/1} & {50.2}\\
        & & {Bi-Real} & {1/1} & {63.8}\\
        & & {BiDet} & {1/1} & {66.0}\\
        & & {DIR-Net} & {1/1} & {\textbf{67.1}$_{\pm0.13}$}\\
        \midrule
        \multirow{5}{*}{{Faster R-CNN (ResNet-18)}} & \multirow{5}{*}{{600$\times$1000}} & {Full-Precision} & {1/1} & {74.5}\\
        & & {BNN} & {1/1} & {35.6}\\
        & & {XNOR} & {1/1} & {48.4}\\
        & & {Bi-Real} & {1/1} & {58.2}\\
        & & {BiDet} & {1/1} & {59.5}\\
        & & {DIR-Net} & {1/1} & {\textbf{60.4}$_{\pm0.07}$}\\
        \bottomrule
    \end{tabular}}}
\end{table}

\subsubsection{Object Detection Tasks}

{\textbf{PASCAL VOC dataset}. Furthermore, to validate the generalization of our DIR-Net, we evaluate it on other vision tasks. PASCAL VOC is a widespread dataset for object detection with 20 classes. We compare our method with many other binarization methods, including BNN~\cite{hubara2016binarized}, XNOR~\cite{xnornet}, Bi-Real~\cite{Liu_2018_ECCV}, and BiDet~\cite{wang2020bidet}. And we also apply binarization on different frameworks, such as SSD300~\cite{liu2016ssd} and Faster R-CNN~\cite{ren2015faster} with VGG-16 and ResNet-18 backbones, respectively.}

{As shown in Table~\ref{pascal-voc}, our DIR-Net far exceeds all existing binarization methods by convincing margin. Compared with the general quantization methods, such as BNN, XNOR, and Bi-Real, our DIR-Net has achieved significant improvement, which is consistent with the phenomenon of classification tasks. 
For the SSD300 and Faster R-CNN frameworks, the DIR-Net outperforms these methods at last 3.3\% and 2.2\% of mAP, respectively.
And compared with BiDet, a binary quantization method specially designed for object detection, DIR-Net improves the performance of 1.1\% (SSD300) and 0.9\% (Faster R-CNN) by improving the training strategy without bringing any inference burden.
For the results of DIR-Net, the standard deviation caused by different random seeds is less than or equal to 0.13\%.}

\begin{table}[htp]
    \caption{Performance comparison on COCO.}
    \label{coco}
    \vspace{-0.1in}
    \centering
    {
    \setlength{\tabcolsep}{2.mm}{
    \begin{tabular}{lcccccc}
        \toprule
        {\tabincell{l}{Framework\\(Backbone)}} & {Input} & {Method} & {\tabincell{c}{Bit-width\\(W/A)}} & {\tabincell{c}{mAP\\$_{@[.5, .95]}$(\%)}} & {\tabincell{c}{AP$_{50}$\\(\%)}} & {\tabincell{c}{AP$_{75}$\\(\%)}} \\
        \midrule
        \multirow{5}{*}{{\tabincell{l}{SSD300\\(VGG-16)}}} & \multirow{5}{*}{{300$\times$300}} & {Full-Precision} & {32/32} & {23.2} & {41.2} & {23.4}\\
        & & {BNN} & {1/1} & {6.2} & {15.9} & {3.8}\\
        & & {XNOR} & {1/1} & {8.1} & {19.5} & {5.6}\\
        & & {Bi-Real} & {1/1} & {11.2} & {26.0} & {8.3}\\
        & & {BiDet} & {1/1} & {13.2} & {28.3} & {10.5}\\
        & & {DIR-Net} & {1/1} & {\textbf{14.0}$_{\pm0.10}$} & {\textbf{29.7}} & {\textbf{11.3}}\\
        \midrule
        \multirow{5}{*}{{\tabincell{l}{Faster R-CNN\\(ResNet-18)}}} & \multirow{5}{*}{{600$\times$1000}} & {Full-Precision} & {1/1} & {26.0} & {44.8} & {27.2}\\
        & & {BNN} & {1/1} & {5.6} & {14.3} & {2.6}\\
        & & {XNOR} & {1/1} & {10.4} & {21.6} & {8.8}\\
        & & {Bi-Real} & {1/1} & {14.4} & {29.0} & {13.4}\\
        & & {BiDet} & {1/1} & {15.7} & {31.0} & {14.4}\\
        & & {DIR-Net} & {1/1} & {\textbf{16.1}$_{\pm0.08}$} & {\textbf{31.5}} & {\textbf{14.8}}\\
        \bottomrule
    \end{tabular}}}
\end{table}

\noindent{\textbf{COCO dataset}.
We also evaluate the DIR-Net on the COCO dataset, which is much more challenging than PASCAL VOC due to the high diversity and large scale.
We compare our method with BNN~\cite{hubara2016binarized}, XNOR~\cite{xnornet}, Bi-Real~\cite{Liu_2018_ECCV}, and BiDet~\cite{wang2020bidet} on different frameworks, including the SSD300~\cite{liu2016ssd} and Faster R-CNN~\cite{ren2015faster} with VGG-16 and ResNet-18 backbones, respectively.}

{As Table~\ref{coco} shows, DIR-Net still far outperforms BNN, XNOR, Bi-Real, and BiDet.
When applying the SSD300 and Faster R-CNN frameworks, DIR-Net outperforms the existing SOTA method BiDet by 0.8\% and 0.4\%, respectively in mAP, and the standard deviation caused by different random seeds is less than or equal to 0.10\%.
Experiments show that the techniques of DIR-Net can stably and significantly improve BNNs on large-scale object detection tasks.}

\subsection{Deployment Efficiency on Raspberry Pi 3B} 
\label{deploy_efficiency}
In order to further evaluate the efficiency of our proposed DIR-Net when it is deployed on real-world mobile devices, we implemented DIR-Net on Raspberry Pi 3B, which has a 1.2 GHz 64-bit quad-core ARM Cortex-A53 and tested the running speed in practice. We use the SIMD instruction SSHL on ARM NEON to ensure the inference framework daBNN~\cite{zhang2019dabnn} is compatible with DIR-Net.

\begin{table}[htp]
    \caption{Comparison of time cost of ResNet-18 with different bits (single thread).}
    \label{table:eff}
    \vspace{-0.1in}
    \centering
    \setlength{\tabcolsep}{4.2mm}{
    \begin{tabular}{lccc}
        \toprule
        Method&\tabincell{c}{Bit-width (W/A)}&\tabincell{c}{Size (Mb)}&\tabincell{c}{Time (ms)}\\
        \midrule
        Full-Precision&32/32& 46.77 & 1418.94 \\
        NCNN&8/8& -- & 935.51 \\
        DSQ&2/2& -- & 551.22 \\
        DIR-Net (w/o scalars)&1/1&\textbf{4.20}&\textbf{252.16}\\
        DIR-Net (ours) &1/1&\textbf{4.21}&\textbf{261.98}\\
        \bottomrule
    \end{tabular}}
    \vspace{-0.1in}
\end{table}

We must point out that so far, very few studies have reported the inference speed of their models deployed on real-world devices which is one of the most important criteria for evaluating the quantized models, especially when using 1-bit binarization. As shown in Table~\ref{table:eff}, we compare DIR-Net with existing high-performance inference implementations including NCNN~\cite{ncnn} and DSQ~\cite{DSQ}. Obviously, the inference speed of DIR-Net is much faster than others since all floating-point operations in convolutional layers are replaced by bitwise operations, such as XNOR, Bitcount, and Bit-shift. And the model size of DIR-Net can also be greatly reduced, the shift-based scalars in DIR-Net bring almost no extra time consumption and memory footprint compared with the vanilla binarization method without scalars.

\section{Conclusion}
In this paper, we introduce a novel DIR-Net that retains the information during the forward/backward propagation of binary neural networks. {The DIR-Net is mainly composed of three practical technologies: the IMB for ensuring diversity in the forward propagation, the DTE for reducing the gradient errors in the backward propagation, and the RBD for retaining the representation information with the help of external representations.} From the perspective of information entropy, IMB performs a simple but effective transformation on weights, which maximizes the information loss of both weights and activations at the same time, with no additional operations on activations. In this way, we can maintain the diversity of binary neural networks as much as possible without compromising efficiency. A well-designed gradient estimator DTE also reduces the information errors of gradients in the backward propagation. Because of the powerful updating capability and accurate gradients, the performance of DTE exceeds that of STE by a large margin. 
{Additionally, with well-trained corresponding full-precision networks, the RBD scheme improves BNNs orthogonally with the improved internal propagation by introducing external representations.}
Our adequate experiments show that DIR-Net consistently outperforms the existing SOTA BNNs. 

\section{Acknowledgement}
This work was supported by The National Key Research and Development Plan of China (2020AAA0103503), National Natural Science Foundation of China (62022009 and 61872021), Beijing Nova Program of Science and Technology (Z191100001119050), and the Academic Excellence Foundation of BUAA for PhD Students.

\clearpage
{
\bibliographystyle{spmpsci}
\bibliography{egbib}

\begin{thebibliography}{100}
\providecommand{\url}[1]{{#1}}
\providecommand{\urlprefix}{URL }
\expandafter\ifx\csname urlstyle\endcsname\relax
  \providecommand{\doi}[1]{DOI~\discretionary{}{}{}#1}\else
  \providecommand{\doi}{DOI~\discretionary{}{}{}\begingroup
  \urlstyle{rm}\Url}\fi

\bibitem{Ajanthan_2019_ICCV}
Ajanthan, T., Dokania, P.K., Hartley, R., Torr, P.H.S.: Proximal mean-field for
  neural network quantization.
\newblock In: IEEE ICCV (2019)

\bibitem{ResNet-20-project}
akamaster: pytorch\_resnet\_cifar10.
\newblock \url{https://github.com/akamaster/pytorch_resnet_cifar10}

\bibitem{ACIQ}
Banner, R., Nahshan, Y., Hoffer, E., Soudry, D.: Post training 4-bit
  quantization of convolution networks for rapid-deployment.
\newblock CoRR \textbf{abs/1810.05723} (2018)

\bibitem{bengio2013estimating}
Bengio, Y., L{\'e}onard, N., Courville, A.: Estimating or propagating gradients
  through stochastic neurons for conditional computation.
\newblock arXiv  (2013)

\bibitem{bethge2021meliusnet}
Bethge, J., Bartz, C., Yang, H., Chen, Y., Meinel, C.: Meliusnet: An improved
  network architecture for binary neural networks.
\newblock In: WACV (2021)

\bibitem{XNOR++}
Bulat, A., Tzimiropoulos, G.: Xnor-net++: Improved binary neural networks.
\newblock CoRR \textbf{abs/1909.13863} (2019)

\bibitem{ImprovedTraining}
Bulat, A., Tzimiropoulos, G., Kossaifi, J., Pantic, M.: Improved training of
  binary networks for human pose estimation and image recognition.
\newblock CoRR \textbf{abs/1904.05868} (2019)

\bibitem{DBLP:conf/cvpr/CaiHSV17}
Cai, Z., He, X., Sun, J., Vasconcelos, N.: Deep learning with low precision by
  half-wave gaussian quantization.
\newblock In: IEEE CVPR (2017)

\bibitem{Cao_2019_CVPR}
Cao, S., Ma, L., Xiao, W., Zhang, C., Liu, Y., Zhang, L., Nie, L., Yang, Z.:
  Seernet: Predicting convolutional neural network feature-map sparsity through
  low-bit quantization.
\newblock In: IEEE CVPR (2019)

\bibitem{chen2020binarized}
Chen, H., Zhang, B., Zheng, X., Liu, J., Ji, R., Doermann, D., Guo, G., et~al.:
  Binarized neural architecture search for efficient object recognition.
\newblock ICCV  (2020)

\bibitem{DBLP:conf/nips/ChenWP19}
Chen, S., Wang, W., Pan, S.J.: Metaquant: Learning to quantize by learning to
  penetrate non-differentiable quantization.
\newblock In: NeurIPS (2019)

\bibitem{chen2018darkrank}
Chen, Y., Zhang, Z., Wang, N.: Darkrank: Accelerating deep metric learning via
  cross sample similarities transfer.
\newblock AAAI  (2018)

\bibitem{DBLP:journals/corr/CourbariauxB16}
Courbariaux, M., Hubara, I., Soudry, D., El-Yaniv, R., Bengio, Y.: Binarized
  neural networks: Training deep neural networks with weights and activations
  constrained to+ 1 or-1.
\newblock CoRR \textbf{abs/1602.02830} (2016)

\bibitem{BNN+}
Darabi, S., Belbahri, M., Courbariaux, M., Nia, V.P.: {BNN+:} improved binary
  network training.
\newblock CoRR \textbf{abs/1812.11800} (2018)

\bibitem{Deng2009ImageNet}
Deng, J., Dong, W., Socher, R., Li, L.J., Li, K., Li, F.F.: Imagenet: a
  large-scale hierarchical image database.
\newblock In: IEEE CVPR (2009)

\bibitem{DBLP:journals/pr/DingCH19}
Ding, H., Chen, K., Huo, Q.: Compressing {CNN-DBLSTM} models for {OCR} with
  teacher-student learning and tucker decomposition.
\newblock Pattern Recognition \textbf{96} (2019)

\bibitem{Regularize-act-distribution}
Ding, R., Chin, T.W., Liu, Z., Marculescu, D.: Regularizing activation
  distribution for training binarized deep networks.
\newblock In: IEEE CVPR (2019)

\bibitem{DBLP:journals/ijcv/DongNLCSZ19}
Dong, Y., Ni, R., Li, J., Chen, Y., Su, H., Zhu, J.: Stochastic quantization
  for learning accurate low-bit deep neural networks.
\newblock Int. J. Comput. Vis. \textbf{127}(11-12), 1629--1642 (2019)

\bibitem{Dong2017Learning}
Dong, Y., Ni, R., Li, J., Chen, Y., Zhu, J., Su, H.: Learning accurate low-bit
  deep neural networks with stochastic quantization.
\newblock BMVC  (2017)

\bibitem{Dong_2019_ICCV}
Dong, Z., Yao, Z., Gholami, A., Mahoney, M.W., Keutzer, K.: Hawq: Hessian aware
  quantization of neural networks with mixed-precision.
\newblock In: IEEE ICCV (2019)

\bibitem{Everingham:2010:PVO:1747084.1747104}
Everingham, M., Van~Gool, L., Williams, C.K.I., Winn, J., Zisserman, A.: The
  pascal visual object classes challenge.
\newblock IJCV

\bibitem{Everingham10}
Everingham, M., Van~Gool, L., Williams, C.K.I., Winn, J., Zisserman, A.: The
  pascal visual object classes (voc) challenge.
\newblock International Journal of Computer Vision \textbf{88}(2), 303--338
  (2010)

\bibitem{ge2017Compressing}
Ge, S., Luo, Z., Zhao, S., Jin, X., Zhang, X.: Compressing deep neural networks
  for efficient visual inference.
\newblock In: IEEE ICME (2017)

\bibitem{DBLP:journals/corr/Girshick15}
Girshick, R.: Fast r-cnn.
\newblock In: IEEE ICCV (2015)

\bibitem{DBLP:journals/corr/GirshickDDM13}
Girshick, R., Donahue, J., Darrell, T., Malik, J.: Rich feature hierarchies for
  accurate object detection and semantic segmentation.
\newblock In: IEEE CVPR (2014)

\bibitem{DSQ}
Gong, R., Liu, X., Jiang, S., Li, T., Hu, P., Lin, J., Yu, F., Yan, J.:
  Differentiable soft quantization: Bridging full-precision and low-bit neural
  networks.
\newblock In: IEEE ICCV (2019)

\bibitem{DBLP:journals/corr/abs-1811-12755}
Gu, J., Li, C., Zhang, B., Han, J., Cao, X., Liu, J., Doermann, D.S.:
  Projection convolutional neural networks for 1-bit cnns via discrete back
  propagation.
\newblock CoRR \textbf{abs/1811.12755} (2018)

\bibitem{gu2019bayesian}
Gu, J., Zhao, J., Jiang, X., Zhang, B., Liu, J., Guo, G., Ji, R.: Bayesian
  optimized 1-bit cnns.
\newblock In: Proceedings of the IEEE/CVF International Conference on Computer
  Vision, pp. 4909--4917 (2019)

\bibitem{han2016deep}
Han, S., Mao, H., Dally, W.J.: Deep compression: Compressing deep neural
  network with pruning, trained quantization and huffman coding.
\newblock ICLR  (2016)

\bibitem{he2016deep}
He, K., Zhang, X., Ren, S., Sun, J.: Deep residual learning for image
  recognition.
\newblock In: IEEE CVPR (2016)

\bibitem{he2021generative}
He, X., Hu, Q., Wang, P., Cheng, J.: Generative zero-shot network quantization.
\newblock arXiv preprint arXiv:2101.08430  (2021)

\bibitem{he2017channel}
He, Y., Zhang, X., Sun, J.: Channel pruning for accelerating very deep neural
  networks.
\newblock In: IEEE ICCV (2017)

\bibitem{Simultaneously-Optimizing-Weight}
He, Z., Fan, D.: Simultaneously optimizing weight and quantizer of ternary
  neural network using truncated gaussian approximation.
\newblock In: IEEE CVPR (2019)

\bibitem{hinton2015distilling}
Hinton, G., Vinyals, O., Dean, J., et~al.: Distilling the knowledge in a neural
  network.
\newblock arXiv preprint arXiv:1503.02531 \textbf{2}(7) (2015)

\bibitem{Loss-Aware-BNN}
Hou, L., Yao, Q., Kwok, J.T.: Loss-aware binarization of deep networks.
\newblock ICLR  (2017)

\bibitem{mobilenet}
Howard, A.G., Zhu, M., Chen, B., Kalenichenko, D., Wang, W., Weyand, T.,
  Andreetto, M., Adam, H.: Mobilenets: Efficient convolutional neural networks
  for mobile vision applications.
\newblock CoRR \textbf{abs/1704.04861} (2017)

\bibitem{DBLP:conf/eccv/HuLWZC18}
Hu, Q., Li, G., Wang, P., Zhang, Y., Cheng, J.: Training binary weight networks
  via semi-binary decomposition.
\newblock In: ECCV (2018)

\bibitem{DBLP:journals/corr/abs-1802-02733}
Hu, Q., Wang, P., ChengT, J.: From hashing to cnns: Training binary weight
  networks via hashing.
\newblock In: AAAI (2018)

\bibitem{hubara2016binarized}
Hubara, I., Courbariaux, M., Soudry, D., El-Yaniv, R., Bengio, Y.: Binarized
  neural networks.
\newblock In: NeurIPS (2016)

\bibitem{jaderberg2014speeding}
Jaderberg, M., Vedaldi, A., Zisserman, A.: Speeding up convolutional neural
  networks with low rank expansions.
\newblock In: BMVC (2014)

\bibitem{Jung_2019_CVPR}
Jung, S., Son, C., Lee, S., Son, J., Han, J.J., Kwak, Y., Hwang, S.J., Choi,
  C.: Learning to quantize deep networks by optimizing quantization intervals
  with task loss.
\newblock In: IEEE CVPR (2019)

\bibitem{35496}
Kamgar-Parsi, B., Kamgar-Parsi, B.: Evaluation of quantization error in
  computer vision.
\newblock IEEE Transactions on Pattern Analysis and Machine Intelligence
  \textbf{11}(9), 929--940 (1989).
\newblock \doi{10.1109/34.35496}

\bibitem{CIFAR}
Krizhevsky, A., Nair, V., Hinton, G.: The cifar-10 dataset.
\newblock online: http://www. cs. toronto. edu/kriz/cifar. html  (2014)

\bibitem{krizhevsky2012imagenet}
Krizhevsky, A., Sutskever, I., Hinton, G.E.: Imagenet classification with deep
  convolutional neural networks.
\newblock In: NeurIPS (2012)

\bibitem{kruger2014benchmarking}
Kruger, C.P., Hancke, G.P.: Benchmarking internet of things devices.
\newblock In: IEEE INDIN, pp. 611--616 (2014)

\bibitem{ResNet-18-project}
kuangliu, ypwhs, fducau, bearpaw: pytorch-cifar.
\newblock \url{https://github.com/kuangliu/pytorch-cifar}

\bibitem{selfBN}
Lahoud, F., Achanta, R., M{\'{a}}rquez{-}Neila, P., S{\"{u}}sstrunk, S.:
  Self-binarizing networks.
\newblock CoRR \textbf{abs/1902.00730} (2019)

\bibitem{lebedev2015speeding}
Lebedev, V., Ganin, Y., Rakhuba, M., Oseledets, I.V., Lempitsky, V.S.:
  Speeding-up convolutional neural networks using fine-tuned cp-decomposition.
\newblock In: ICLR (2015)

\bibitem{lebedev2016fast}
Lebedev, V., Lempitsky, V.: Fast convnets using group-wise brain damage.
\newblock In: IEEE CVPR (2016)

\bibitem{DBLP:journals/corr/LiL16}
Li, F., Zhang, B., Liu, B.: Ternary weight networks.
\newblock CoRR \textbf{abs/1605.04711} (2016)

\bibitem{Li_2019_CVPR}
Li, R., Wang, Y., Liang, F., Qin, H., Yan, J., Fan, R.: Fully quantized network
  for object detection.
\newblock In: IEEE CVPR (2019)

\bibitem{DBLP:journals/corr/abs-1708-08687}
Li, Z., Ni, B., Zhang, W., Yang, X., Gao, W.: Performance guaranteed network
  acceleration via high-order residual quantization.
\newblock In: IEEE ICCV (2017)

\bibitem{lin2018defensive}
Lin, J., Gan, C., Han, S.: Defensive quantization: When efficiency meets
  robustness.
\newblock In: International Conference on Learning Representations (2019)

\bibitem{lin2014microsoft}
Lin, T.Y., Maire, M., Belongie, S., Hays, J., Perona, P., Ramanan, D.,
  Doll{\'a}r, P., Zitnick, C.L.: Microsoft coco: Common objects in context.
\newblock In: European conference on computer vision, pp. 740--755. Springer
  (2014)

\bibitem{ABCNet}
Lin, X., Zhao, C., Pan, W.: Towards accurate binary convolutional neural
  network.
\newblock In: NeurIPS (2017)

\bibitem{DBLP:journals/ijcv/LiuDHZLGD21}
Liu, C., Ding, W., Hu, Y., Zhang, B., Liu, J., Guo, G., Doermann, D.S.:
  Rectified binary convolutional networks with generative adversarial learning.
\newblock Int. J. Comput. Vis. \textbf{129}(4), 998--1012 (2021)

\bibitem{Liu_2019_CVPR}
Liu, C., Ding, W., Xia, X., Zhang, B., Gu, J., Liu, J., Ji, R., Doermann, D.:
  Circulant binary convolutional networks: Enhancing the performance of 1-bit
  dcnns with circulant back propagation.
\newblock In: CVPR (2019)

\bibitem{liu2018darts}
Liu, H., Simonyan, K., Yang, Y.: Darts: Differentiable architecture search.
\newblock arXiv preprint arXiv:1806.09055  (2018)

\bibitem{liu2016ssd}
Liu, W., Anguelov, D., Erhan, D., Szegedy, C., Reed, S., Fu, C.Y., Berg, A.C.:
  Ssd: Single shot multibox detector.
\newblock In: European conference on computer vision, pp. 21--37. Springer
  (2016)

\bibitem{DBLP:journals/ijcv/LiuLWYLC20}
Liu, Z., Luo, W., Wu, B., Yang, X., Liu, W., Cheng, K.: Bi-real net: Binarizing
  deep network towards real-network performance.
\newblock Int. J. Comput. Vis. \textbf{128}(1), 202--219 (2020)

\bibitem{liu2020reactnet}
Liu, Z., Shen, Z., Savvides, M., Cheng, K.T.: Reactnet: Towards precise binary
  neural network with generalized activation functions.
\newblock In: ECCV (2020)

\bibitem{Liu_2018_ECCV}
Liu, Z., Wu, B., Luo, W., Yang, X., Liu, W., Cheng, K.T.: Bi-real net:
  Enhancing the performance of 1-bit cnns with improved representational
  capability and advanced training algorithm.
\newblock In: ECCV (2018)

\bibitem{loshchilov2016sgdr}
Loshchilov, I., Hutter, F.: Sgdr: Stochastic gradient descent with warm
  restarts.
\newblock arXiv preprint arXiv:1608.03983  (2016)

\bibitem{martinez2020training}
Martinez, B., Yang, J., Bulat, A., Tzimiropoulos, G.: Training binary neural
  networks with real-to-binary convolutions.
\newblock In: ICLR (2020)

\bibitem{mishra2018wrpn}
Mishra, A., Nurvitadhi, E., Cook, J.J., Marr, D.: {WRPN}: Wide
  reduced-precision networks.
\newblock In: ICLR (2018)

\bibitem{Morozov_2019_ICCV}
Morozov, S., Babenko, A.: Unsupervised neural quantization for
  compressed-domain similarity search.
\newblock In: IEEE ICCV (2019)

\bibitem{Nagel_2019_ICCV}
Nagel, M., Baalen, M.v., Blankevoort, T., Welling, M.: Data-free quantization
  through weight equalization and bias correction.
\newblock In: IEEE ICCV (2019)

\bibitem{ncnn}
nihui, BUG1989, Howave, gemfield, Corea, eric612: ncnn.
\newblock \url{https://github.com/Tencent/ncnn}

\bibitem{DBLP:journals/corr/abs-1904-02701}
Pang, J., Chen, K., Shi, J., Feng, H., Ouyang, W., Lin, D.: Libra r-cnn:
  Towards balanced learning for object detection.
\newblock In: IEEE CVPR (2019)

\bibitem{phan2020binarizing}
Phan, H., Liu, Z., Huynh, D., Savvides, M., Cheng, K.T., Shen, Z.: Binarizing
  mobilenet via evolution-based searching.
\newblock In: CVPR (2020)

\bibitem{qin2020bipointnet}
Qin, H., Cai, Z., Zhang, M., Ding, Y., Zhao, H., Yi, S., Liu, X., Su, H.:
  Bipointnet: Binary neural network for point clouds.
\newblock arXiv preprint arXiv:2010.05501  (2020)

\bibitem{IRNet}
Qin, H., Gong, R., Liu, X., Shen, M., Wei, Z., Yu, F., Song, J.: Forward and
  backward information retention for accurate binary neural networks.
\newblock In: CVPR (2020)

\bibitem{xnornet}
Rastegari, M., Ordonez, V., Redmon, J., Farhadi, A.: Xnor-net: Imagenet
  classification using binary convolutional neural networks.
\newblock In: ECCV (2016)

\bibitem{NIPS2015_5638}
Ren, S., He, K., Girshick, R., Sun, J.: Faster r-cnn: Towards real-time object
  detection with region proposal networks.
\newblock In: NeurIPS (2015)

\bibitem{ren2015faster}
Ren, S., He, K., Girshick, R., Sun, J.: Faster r-cnn: Towards real-time object
  detection with region proposal networks.
\newblock Advances in neural information processing systems \textbf{28} (2015)

\bibitem{VeryDeepConvolutional}
Simonyan, K., Zisserman, A.: Very deep convolutional networks for large-scale
  image recognition.
\newblock arXiv preprint arXiv:1409.1556  (2014)

\bibitem{DBLP:journals/ijcv/SongHGXHS20}
Song, J., He, T., Gao, L., Xu, X., Hanjalic, A., Shen, H.T.: Unified binary
  generative adversarial network for image retrieval and compression.
\newblock Int. J. Comput. Vis. \textbf{128}(8), 2243--2264 (2020)

\bibitem{7298594}
Szegedy, C., Liu, W., Jia, Y., Sermanet, P., Reed, S., Anguelov, D., Erhan, D.,
  Vanhoucke, V., Rabinovich, A.: Going deeper with convolutions.
\newblock In: IEEE CVPR (2015)

\bibitem{tan2019efficientnet}
Tan, M., Le, Q.: Efficientnet: Rethinking model scaling for convolutional
  neural networks.
\newblock In: ICML (2019)

\bibitem{Wang_2019_CVPR}
Wang, K., Liu, Z., Lin, Y., Lin, J., Han, S.: Haq: Hardware-aware automated
  quantization with mixed precision.
\newblock In: IEEE CVPR (2019)

\bibitem{wang2020towards}
Wang, P., Chen, Q., He, X., Cheng, J.: Towards accurate post-training network
  quantization via bit-split and stitching.
\newblock In: ICML (2020)

\bibitem{wang2020sparsity}
Wang, P., He, X., Li, G., Zhao, T., Cheng, J.: Sparsity-inducing binarized
  neural networks.
\newblock In: AAAI (2020)

\bibitem{wang2019dynamic}
Wang, Y., Gan, W., Wu, W., Yan, J.: Dynamic curriculum learning for imbalanced
  data classification.
\newblock In: IEEE ICCV (2019)

\bibitem{DBLP:journals/pami/WangXXT19}
Wang, Y., Xu, C., Xu, C., Tao, D.: Packing convolutional neural networks in the
  frequency domain.
\newblock IEEE TPAMI  (2019)

\bibitem{wang2020bidet}
Wang, Z., Wu, Z., Lu, J., Zhou, J.: Bidet: An efficient binarized object
  detector.
\newblock In: Proceedings of the IEEE/CVF Conference on Computer Vision and
  Pattern Recognition, pp. 2049--2058 (2020)

\bibitem{DBLP:journals/pr/WenZXYH18}
Wen, J., Zhang, B., Xu, Y., Yang, J., Han, N.: Adaptive weighted nonnegative
  low-rank representation.
\newblock Pattern Recognition \textbf{81}, 326--340 (2018)

\bibitem{93812}
Wong, P.: On quantization errors in computer vision.
\newblock IEEE Transactions on Pattern Analysis and Machine Intelligence
  \textbf{13}(9), 951--956 (1991).
\newblock \doi{10.1109/34.93812}

\bibitem{9027877}
{Wu}, Y., {Liu}, X., {Qin}, H., {Xia}, K., {Hu}, S., {Ma}, Y., {Wang}, M.:
  Boosting temporal binary coding for large-scale video search.
\newblock IEEE Transactions on Multimedia pp. 1--1 (2020)

\bibitem{Wu2020Rotation}
Wu, Y., Wu, Y., Gong, R., Lv, Y., Chen, K., Liang, D., Hu, X., Liu, X., Yan,
  J.: Rotation consistent margin loss for efficient low-bit face recognition.
\newblock CoRR  (2020)

\bibitem{diverse}
Xie, B., Liang, Y., Song, L.: Diverse neural network learns true target
  functions.
\newblock In: Artificial Intelligence and Statistics (2017)

\bibitem{Yang_2019_CVPR}
Yang, J., Shen, X., Xing, J., Tian, X., Li, H., Deng, B., Huang, J., Hua, X.s.:
  Quantization networks.
\newblock In: IEEE CVPR (2019)

\bibitem{yu2017on}
Yu, X., Liu, T., Wang, X., Tao, D.: On compressing deep models by low rank and
  sparse decomposition.
\newblock In: IEEE CVPR (2017)

\bibitem{zagoruyko2017paying}
Zagoruyko, S., Komodakis, N.: Paying more attention to attention: Improving the
  performance of convolutional neural networks via attention transfer.
\newblock In: ICLR (2017)

\bibitem{LQ-Net}
Zhang, D., Yang, J., Ye, D., Hua, G.: Lq-nets: Learned quantization for highly
  accurate and compact deep neural networks.
\newblock In: ECCV (2018)

\bibitem{zhang2019dabnn}
Zhang, J., Pan, Y., Yao, T., Zhao, H., Mei, T.: dabnn: {A} super fast inference
  framework for binary neural networks on {ARM} devices.
\newblock In: ACM MM (2019)

\bibitem{shufflenet}
Zhang, X., Zhou, X., Lin, M., Sun, J.: Shufflenet: An extremely efficient
  convolutional neural network for mobile devices.
\newblock In: IEEE CVPR (2018)

\bibitem{dorefa}
Zhou, S., Wu, Y., Ni, Z., Zhou, X., Wen, H., Zou, Y.: Dorefa-net: Training low
  bitwidth convolutional neural networks with low bitwidth gradients.
\newblock CoRR \textbf{abs/1606.06160} (2016)

\bibitem{DBLP:journals/corr/ZhuHMD16}
Zhu, C., Han, S., Mao, H., Dally, W.J.: Trained ternary quantization.
\newblock In: ICLR (2017)

\bibitem{zhu2019unified}
Zhu, F., Gong, R., Yu, F., Liu, X., Wang, Y., Li, Z., Yang, X., Yan, J.:
  Towards unified int8 training for convolutional neural network (2019)

\bibitem{Zhuang_2019_CVPR}
Zhuang, B., Shen, C., Tan, M., Liu, L., Reid, I.: Structured binary neural
  networks for accurate image classification and semantic segmentation.
\newblock In: IEEE CVPR (2019)

\end{thebibliography}
}

\end{document}